# A novel approach to the relationships between data features
## - based on comprehensive examination of mathematical, technological, and causal methodology

JaeHong Kim[1, 2]

[1] Legal Research Institute of Korea University, Seoul 02841, Korea

[2] Human-Inspired AI Research, Korea University, Seoul 02841, Korea

Contact email: comforta@korea.ac.kr

# Abstract

The expansion of artificial intelligence (AI) has raised concerns about transparency, accountability, and interpretability, with counterfactual reasoning emerging as a key approach to addressing these issues. However, current mathematical, technological, and causal methodologies rely on externalization techniques that normalize feature relationships within a single coordinate space, often distorting intrinsic interactions. This study proposes the Convergent Fusion Paradigm (CFP) theory, a framework integrating mathematical, technological, and causal perspectives to provide a more precise and comprehensive analysis of feature relationships. CFP theory introduces Hilbert space and backward causation to reinterpret the feature relationships as emergent structures, offering a potential solution to the common cause problem—a fundamental challenge in causal modeling. From a mathematical-technical perspective, it utilizes a Riemannian manifold-based framework, thereby improving the structural representation of high- and low-dimensional data interactions. From a causal inference perspective, CFP theory adopts abduction as a methodological foundation, employing Hilbert space for a dynamic causal reasoning approach, where causal relationships are inferred abductively, and feature relationships evolve as emergent properties. Ultimately, CFP theory introduces a novel AI modeling methodology that integrates Hilbert space, backward causation, and Riemannian geometry, strengthening AI governance and transparency in counterfactual reasoning.

# I. Introduction

The issues of transparency and accountability in artificial intelligence (AI) represent some of the most critical legal and ethical challenges arising from the pervasive integration of AI into modern society as a leading core technology of the 21st century. [To address these challenges,] there has been much discussion of the existence of a "right to explanation" in the EU General Data Protection Regulation (GDPR), and its merits and disadvantages (Edwards and Veale 2017; Malgieri and Comandé 2017; Mendoza and Bygrave 2017; Selbst and Powles 2018; Wachter, Mittelstadt, and Floridi 2017; Wachter, Mittelstadt, and Russell 2017, p842)

However, explanations provided to data subjects do not necessarily require them to fully understand the operational mechanisms of algorithmic systems. Instead, revealing the system's inner workings could infringe on the rights of counterparties, such as trade secrets and privacy, or increase the likelihood of data subjects manipulating or gaming the decision-making system. Therefore, explanations should aim solely to assist data subjects, as this approach can resolve the issues of algorithmic transparency and accountability without undermining trust between data subjects and data controllers. In principle, such explanations should be provided without disclosing the "black box" of the algorithm. (Wachter, Mittelstadt, and Russell 2017) argue that the concept of "unconditional counterfactual explanation(counterfactuals)" meets these requirements and overcomes many challenges related to algorithmic interpretability and accountability. Counterfactuals have been discussed as a key technique within the theme of algorithmic explainability, which is central to explainable Artificial Intelligence (xAI) (Molnar 2020).

The process of generating counterfactuals involves two steps: first, identifying a decision that contradicts the data subject's intent; second, determining the minimal changes required for the decision model to produce a favorable outcome for the data subject. Notably, these changes are defined at the feature level of the data. For instance, in a "loan approval/denial decision model," if a loan application is denied based on the submitted data, the features requiring modification to secure approval may include *employment period, job change, salary,* and *bank account balance.* Understanding the relationships among features like *employment period, salary,* and *bank account balance* is essential to providing meaningful responses to data subjects through counterfactuals.

To better understand the relationships among data features, the following questions may arise:

Is there a direct or inverse relationship between *employment period* and *job change*?

To increase *salary*, should *employment period* be extended, or is a *job change* more appropriate? Could *salary* be considered a common cause of both *employment period* and *job change*?

How does a change in *salary* affect the *bank account balance*? Could *salary* be viewed as a higher-level feature of *bank account balance*?

As these questions illustrate, the crucial challenge in generating counterfactuals lies in accurately capturing the relationships among data features. However, properly identifying and structuring these relationships is a highly complex and challenging task, involving intertwined technical, mathematical, logical, and philosophical issues. Nevertheless, if AI, as a data science, fails to go beyond the mathematical and technical constraints of data processing and represent the real-world relationships among features accurately, a series of legal (right to explanation) and theoretical (xAI) efforts to ensure AI transparency and accountability will inevitably remain incomplete. Moreover, these feature relationships align closely with the limitations of current mathematical, technical, and causal methodologies. Thus, efforts to rigorously elucidate these relationships are one of the key prerequisites for addressing the broader social and legal issues associated with AI.

This study aims to clarify the relationships among data features by identifying their inherent limitations and proposing a series of strategies to overcome them. Also, it approaches multidisciplinary by integrating the mathematical, technological, and causal perspectives. To this end, the study reviews the noteworthy literature on feature relationships within the past decade of AI and computer science, particularly those related to counterfactuals. It highlights the methodological limitations underlying these studies and proposes the Convergent Fusion Paradigm (CFP) theory (Kim and Shim 2025) as a philosophical and theoretical foundation for overcoming these challenges.

The CFP theory, from the mathematical and technological perspective, particularly focusing on its integrative potential to unify these domains for a cohesive framework, provides new insights into the geometric structure of data related to Riemannian manifolds, offering a structural framework for converging high- and low-dimensional data, which facilitates a clearer understanding of the data generation process. Moreover, this study argues that the "common cause" problem, often regarded as a logical conundrum in applying causal methodologies, could be naturally understood as emergent features by extending the limited spatial or temporal coordinates (Buckner 2020) such as Euclidean space into Hilbert space and equipping it with backward causation. In this context, the CFP theory offers a methodological insight into establishing a new causal model employing emergent feature. Lastly, in this context, this study further posits that abduction could serve as a basis for a new causal model in Hilbert space.

To achieve the above objectives, this study proceeds as follows.

**In Sect. 2**, this study reviews the noteworthy AI literature related to counterfactuals, focusing on how the relationships among features are addressed. The literature is categorized based on whether it is dealt in monotonic geometry or multiple geometry, with representative works from each approach being summarized. By doing so, this study identifies the characteristics and limitations of the methodologies employed in these works. Additionally, this literature is classified according to externalization and internalization methods, as well as mathematical-technological and causal methodologies. Particular attention is given to the limitations posed by the "common cause" problem when addressing feature relationships from a causal perspective. To address this issue as a problem of emergent features, this study emphasizes the need for new conditions and causal reasoning. Subsequently, it examines Hilbert space and backward causation as a framework for space and causal logic, arguing for the necessity of a theoretical foundation (CFP theory) to implement a new causal model aligned with the properties of Hilbert space.

**Sect. 3** begins from a more fundamental perspective to establish the CFP theory. Since AI research is closely linked to data statistics, this study briefly examines the two cultures of statistical modeling—data modeling culture and algorithmic modeling culture—proposed by (Breiman 2001). Following this, this study defines the concept of "data-algorithm relationship modeling" as the unique statistical modeling approach of Deep Neural Networks (DNNs). The analysis highlights the differences between data statistics and AI, particularly DNNs, and conceptualizes the developmental phase created by the relationship modeling of data and algorithms as an "expansion of dimensions accompanied by qualitative transformation." To illustrate this developmental phase, two conceptual hypotheses are proposed. These are informed by achievements in physics and biology, as well as Hofstadter's reformulation of Gödel's proof into a new formula, $(\text{TNT} + G_\omega - \text{PROOF} - \text{PAIR}\{a, a'\})$ (Hofstadter 1999), culminating in the proposals of "creating relative Space-Time in Relationship (crSTR)" and "Duplex Contradictory Paradoxical Stratified structure of Thorough Closure-Eternal Opening (DCPSs of TC-EO)."

**Sect. 4** applies the CFP theory established in Sect. 3 to both mathematical-technological and causal perspectives. First, from the mathematical-technological perspective, it establishes the philosophical and theoretical insights gained when reviewing the works of (Joshi et al. 2019) and (Arvanitidis et al. 2018, 2020) through the lens of the CFP theory. In sequence, from the causal perspective, this study applies the framework of the CFP theory to the logic step of abduction, providing a detailed outline of a methodological foundation for a new causal model addressing emergent features.

**Finally**, the conclusion explores the potential of Hilbert space in data generation and examines why philosophical reflection is essential to support this endeavor.

## II. Theoretical and Philosophical Challenges Related to Data Feature Relationships

As mentioned earlier, it is essential to review how the relationships among data features are addressed in the existing AI literature, particularly those related to counterfactuals. Such a review allows for a comprehensive understanding of the current mathematical, technological, and logical approaches to data feature relationships. Moreover, it provides insights into the theoretical and philosophical underpinnings of these approaches while clarifying their limitations.

### 1. Overview of Existing AI Literature on Data Feature Relationships

#### 1) Methods of Understanding Relationships Among Features
#### – Introduction of Distance Concepts

Most existing AI literature interprets the relationships among data features in terms of the concept of distance between various observed feature values and external reference points, such as decision boundaries (Ustun et al. 2019) or population mean values (Gupta et al. 2019). To use distance as a basis, features must be normalized into certain reference feature values. These normalization processes ensure that features are on a common scale, enabling meaningful comparisons. For instance, the increases in employment period and salary do not naturally correspond to each other. In such cases, normalization attempts to capture the fact that salaries may vary on the order of thousands or tens of thousands of dollars, but length of employment (in years, say) varies at a numerically much smaller scale (Barocas et al. 2020, p83). In other words, normalization assigns appropriate weights to features, allowing them to be compared mathematically within a unified coordinate plane (non-dimensionalize). Typically, this normalization process is calculated using the distance metric of the features.

Additionally, AI literature addressing the relationships among data features can be classified based on whether it employs monotonic geometry (simple geometric space) or multiple geometry (complex geometric space). This study categorizes the former as an "Externalization Methodology (EM)" for understanding feature relationships and the latter as an "Internalization Methodology (IM)".

## 2) Monotonic Geometry (MG) – Externalization Methodology (EM)

Representative works employing externalization methodology include (Ustun et al. 2019) and (Wachter, Mittelstadt, and Russell 2017).

### A. (Wachter, Mittelstadt, and Russell 2017)

Wachter's study was the first to apply counterfactuals to concrete cases, namely the LSAT Dataset and Pima Diabetes Database, in order to derive solutions. Nevertheless, this paper did not establish a decision boundary and normalize the distances between features to compute the counterfactuals. Instead, it formulated the relationships between features through one-hot encoding (Verma et al. 2020) and then computed counterfactuals by normalizing both the original condition world and the counterfactual condition world with respect to the original data by means of a distance metric.

$$\arg\min_{\omega}[l(f_\omega(x_i), y_i) + \rho(\omega)], \tag{1}$$

$$\arg\min_{x'}\min_{\lambda} \lambda(f_\omega(x') - y')^2 + d(x_i, x'), \tag{2}$$

where the label is $y_i$, data point $x_i$ and $\rho(\omega)$ is a regularizer over the weights. To find counterfactual data point $x'$ as close to the original data $x_i$ as possible such that $f_\omega(x')$ is equal to new target $y'$.

To elaborate, in the LSAT Database, data features such as Race (Black or White), GPA, LSAT, and Score inherently differ in scale and dimensionality. However, these features possess characteristics of discrete variables, allowing them to be encoded as binary or categorical values through one-hot encoding (Verma et al. 2020). This encoding process ensures that each feature can be compared within a single coordinate plane. After implementing this encoding, Wachter established a decision boundary using Equation (1) with a weight $\omega$ based on the original data. Then, Equation (2) was employed to derive counterfactuals based on the original data world, applying a normalization process using both L1 and L2 norms, when it comes to the LSAT database. This process facilitated the relative adjustment of feature interactions based on quantitative magnitude (Eq. (3)). Consequently, the counterfactual outcomes derived through this methodology became comparable to the original data within a single coordinate plane, allowing the paper to reveal racial discrimination between Black and White applicants in law school admissions.

This stripped-down version of the LSAT dataset is used in the fairness literature, as classifiers trained on this data naturally exhibit bias against 'black' people (Kusner et al. 2017; Russell et al. 2017). As a result, we will find evidence of this bias in our neural network in some of the counterfactuals we generate (Wachter, Mittelstadt, and Russell 2017, p856).

$$d(x_i, x') = \sum_{k \in F} \frac{|x_{i,k} - x_k'|}{MA\ k}, \qquad (3)$$

where $d(x_i, x')$ is distance between original data point $x_i$ and counterfactual data point $x_k'$, and $MAD_k$ stands for the Manhattan distance metric (L1 norm) which is $MAD_k = \text{median}_{(j \in P)}(|X_{(j,k)} - \text{median}_{(l \in P)}(X_{(l,k)})|)$ for the median absolute deviation of feature $k$, over the set of points $P$.

$$d(x_i, x') = \sum_{k \in F}(x_{i,k} - x_k')^2, \qquad (4)$$

where $d(x_i, x')$ is set of distance of unweighted squared Euclidean distance (L2 norm).

However, this series of processes deviates from the fundamental intent of counterfactual reasoning. Counterfactuals are originally intended to determine the minimum change in a cause necessary to alter the outcome within a causal framework. However, in the LSAT database as analyzed by (Wachter, Mittelstadt, and Russell 2017), the decision boundary, which determines the outcome change, is not explicitly defined. In this situation, (Wachter, Mittelstadt, and Russell 2017) set a single causal variable (the score) to '0' and used the corresponding changes in other causal variables as counterfactuals. Yet, as (Wachter, Mittelstadt, and Russell 2017) themselves acknowledge in their paper, there is no guarantee that the "synthetic counterfactual $x'$" corresponds to a valid data point.

To demonstrate the importance of the choice of distance function, we illustrate below the impact of varying $d(\cdot,\cdot)$ on the LSAT dataset. A further challenge lies in ensuring that the synthetic counterfactual $x'$ corresponds to a valid data point (Wachter, Mittelstadt, and Russell 2017, p855).

This is a misinterpretation arising from the limitations of mathematical properties—first, there is no guarantee that a sufficiently close solution aligns with the decision boundary. If the decision boundary is determined using pre-change conditions while the premise itself is altered, an inconsistency between the already-applied and altered assumption arises. In other words, it is impossible to represent both the original data condition world and the counterfactual data condition world on a single, unified scale.

Put differently, counterfactual reasoning should determine the necessary changes in features (causes) to alter an outcome—for example, what changes in features would enable Tom, who was rejected from law school, to be admitted? Instead, (Wachter, Mittelstadt, and Russell 2017)'s method fixes one causal variable and examines the relative changes in other causal variables, representing them within a single coordinate plane based on the original data. This approach is inherently impossible, because although both the original data world and the counterfactual world may share seemingly identical attributes (such as race, LSAT, and score), they fundamentally belong to distinct worlds that cannot be represented within the same coordinate system. In summary, the counterfactual approach of (Wachter, Mittelstadt, and Russell 2017) merely fixes one causal factor while comparing the relationships among others, allowing for the sociological interpretation such as racial bias. However, this methodology deviates significantly from the original intent of the counterfactual method, which seeks to infer causal changes necessary for an outcome to be altered.

**B. (Ustun et al. 2019)**

(Ustun et al. 2019) established a decision boundary and resolved the relationships among features by computing distances between them. In their paper, they introduced an approach that calculates the optimal combination of features that allow an individual to cross the decision boundary, generating a flipset as a recommendation. The decision subject then selects the most appropriate combination from this flipset. In other words, (Ustun et al. 2019) devised a mathematical formulation to generate flipset and implemented an algorithm capable of computing them.

**Algorithm 1** Enumerate $T$ Minimal Cost Actions for Flipset

**Input**
- IP     instance of (2) for coefficients $w$, features $x$, and actions $A(x)$
- $T \geq 1$     number of items in flipset

**Initialize**
- $\mathcal{A} \leftarrow \emptyset$     actions shown in flipset

1: **repeat**
2:     $a^* \leftarrow$ optimal solution to IP
3:     $\mathcal{A} \leftarrow \mathcal{A} \cup \{a^*\}$     add $a^*$ to set of optimal actions
4:     $S \leftarrow \{j : a_j^* \neq 0\}$     indices of features altered by $a^*$
5:     add constraint to IP to remove actions that alter features $j \in S$:
$$\sum_{j \in S}(1 - u_j) + \sum_{j \notin S} u_j \leq d - 1.$$
6: **until** $|\mathcal{A}| = T$ or IP is infeasible

**Output:** $\mathcal{A}$     actions shown in flipset

Figure 1. Algorithmic approach of cost action formula consists flipset. (Image is adapted from (Ustun et al. 2019))

**Methodological Overview**

Notably, (Ustun et al. 2019) introduced the concept of recourse, which differs from the counterfactual approach proposed by (Wachter, Mittelstadt, and Russell 2017), and designed a methodology that identifies actionable features that can be modified. For example, if an individual is denied a credit loan, the actionable features that could be adjusted include reducing debt or increasing savings. The set of these recommended features is referred to as the flipset, wherein immutable features, such as age and gender, are excluded.

To elaborate, the series of calculations to derive the optimal combination mentioned above is conducted within a linear integer programming framework. The entire process revolves around identifying an action set that minimizes a cost function[1]. Here, minimizing the cost function refers to achieving minimal changes in feature values that result in minimal changes in function values. During this minimization process, constraints are imposed to ensure that only actionable features are considered, while immutable features or those mutable in an infeasible way are excluded. In summary, (Ustun et al. 2019) formulated an approach that identifies the feature values minimizing the cost function, iteratively adjusting them to discover a group of flipset capable of crossing the decision boundary, which the decision subject then selects.

However, this methodology presents fundamental limitations in establishing relationships among features.

**Fundamental Limitations of This Approach**

**a. Arbitrary Nature of Normalization in Decision-Making**

[From the perspective of the decision-maker, the] normalization [process] based simply on the distribution of data is somewhat arbitrary. [For instance,] one decision maker might scale the axes such that increasing income by $5,000 annually is equivalent to an additional year on the job. A competing lender, using different training data, could conclude that $10,000 of income corresponds to one year of work. These lenders might therefore produce different explanations depending on the scaling of attributes (Barocas et al. 2020, p84). While differences in training datasets used by the two cases may explain these discrepancies, they do not justify such arbitrariness.

Normalization techniques typically rely entirely on data distribution and do not incorporate an external point of reference. However, without an external point of reference to ground these scales, the meaning of the relative difference in feature values is unclear (Barocas et al. 2020). In practice, decision-makers consider an external decision boundary to determine the optimal combination that crosses it, making normalization indispensable. As a result, the relationships among features are no longer determined by their intrinsic interactions but rather dictated by external objectives, such as external points of reference or an external decision boundary. This process distorts the genuine nature of feature relationships, as they become subject to external constraints rather than reflecting inherent dependencies. In our study, the normalization process, which evaluates feature relationships solely through quantitative weighting within a single coordinate plane, is referred to as externalization. This externalization process is akin to sketching an object labeled "relationship" on a canvas—a mere approximation rather than a faithful representation of its true structure.

---

[1] $\min \text{cost}(a; x)$ s.t $f(x + a) = 1$ and $a \in A(x)$. Where $A(x) \to \mathbb{R}^+$ is a cost function that encodes preferences between actions, or measures quantities of interests for an audit. Users can specify any cost function that satisfies two properties: (i) $\text{cost}(0; x) = 0$ (no action ↔ no cost); (ii) $\text{cost}(a; x) \leq \text{cost}(a + \epsilon 1_j; x)$ (larger actions ↔ higher cost) (Ustun et al. 2019, p12).

**b. Overly Simplistic and Coarse Representation of Feature Relationships**

Another significant limitation of the externalization process is that it oversimplifies and coarsely represents the intricate relationships among features. To better understand this, we need to first recognize an implicit assumption in (Ustun et al. 2019). An implicit assumption in formulations of recourse is that if attribute $X$ is actionable and attribute $Y$ is actionable, then $X$ and $Y$ are jointly actionable. We contend that this is not necessarily so (Venkatasubramanian and Alfano 2020, p290).

Consider a case where an individual seeks a loan and can modify employment period, income, or job change as actionable features. Suppose the decision subject can either increase their salary by $3,000 or extend their employment period by eight months to improve their creditworthiness.

○ If the individual switches jobs, their income increases by $2,000 immediately. In this case, switching jobs directly results in a $2,000 salary increase.

○ However, job changes also reset employment period, meaning the individual's employment history is negatively impacted.

This demonstrates that changing a single feature (job change) can positively affect one feature (income) while negatively impacting another (employment period). Furthermore, the magnitude of these effects differs across features. Yet, flipset methodology (Ustun et al. 2019)—relying solely on quantitative weighting *via* normalization—fails to capture these complex interdependencies among features.

**3) Multiple Geometry (MG) – Internalization Methodology (IM)**

Representative works addressing the mathematical, technological, and logical challenges mentioned above include (Arvanitidis et al. 2018, 2019, 2020, 2021; Dombrowski et al. 2024; Dominguez-Olmedo et al. 2023; Joshi et al. 2019; Karimi, von Kügelgen, et al. 2020; Karimi et al. 2021, 2023; Kügelgen et al. 2023; Louizos et al. 2017; Pegios et al. 2024). These works avoid reducing the relationships among features to external criteria (externalization process) and instead incorporate the relationship between topological phase of features. This approach is referred to as the Internalization Methodology (IM). This study focuses particularly on the works of (Arvanitidis et al. 2018, 2020, 2021; Dombrowski et al. 2024; Dominguez-Olmedo et al. 2023; Joshi et al. 2019; Pegios et al. 2024).

## A. (Joshi et al. 2019)

(Joshi et al. 2019) [modifies, as a purely mathematical approach to addressing this,] the underlying geometry in which the distance associated with recourse is being calculated. In particular, [the] dependency among attributes correspond[s] to moving along a submanifold of the underlying vector space – informally, [it is] a curved surface in the space, rather than the entirety of the space itself (Venkatasubramanian and Alfano 2020, p292). By adopting this method, the distance to the boundary of the manifold becomes the key to understanding feature relationships. Unlike (Ustun et al. 2019), which measure distances non-differentially in a two-dimensional coordinate plane based on external conditions (e.g., decision boundaries), (Joshi et al. 2019) calculate distances along submanifolds implied in the coordinate plane, thus revealing the topology of feature relationships.

(Joshi et al. 2019) addresses this issue as an optimization problem through the following equation.

$$x' = \arg \min_{z \sim G_\theta(z)} \min_\lambda l\left(\left(\log y_{do(t)}, 1\right) + \lambda c(x^*, G_\theta(z))\right) \tag{1}$$

However, it is crucial to note that even though the distance calculation method of (Joshi et al. 2019) is conducted under the assumption of the manifold hypothesis, it must also be examined through the lens of the practical implementation of dimensional expansions in Euclidean space. From this perspective, the submanifold induced by optimization method (Joshi et al. 2019) creates a virtual surface that differs from the actual representation of dimensional expansion. This is because the optimization process (Joshi et al. 2019) method is performed along external boundaries, rather than in the intrinsic space of the expanding dimensions (manifold) — [a point also highlighted as] informally, a curved surface in the space, rather than the entirety of the space itself (Venkatasubramanian and Alfano 2020, p292). Therefore, the optimization performed along external boundaries may be associated with the reality of dimensional expansion but ultimately creates a virtual submanifold that is somewhat aloof from the actual dimensional expansion itself.

In consideration of the above aspect, describing the dependence of features on a virtual surface following the external boundaries of all dimensions (Venkatasubramanian and Alfano 2020) essentially separates the understanding of dimensional expansion (represented as a reduction in the coordinate plane) from feature relationships. While dimensional expansion is modeled through the submanifold, feature relationships are instead processed as properties of a virtual surface. Such a separation, even when premised on Euclidean space, contradicts the fundamental requirement that features must exist, at least, within the dimensional space itself. Consequently, while the primary advantage of (Joshi et al. 2019)—revealing the topology of feature relationships through distance calculations based on a submanifold—is critical, it is, to a degree, diluted due to the separation between dimensional expansion and feature relationships.

Equally important as addressing the mathematical challenges is tackling the logical and technical tasks of encoding these manifolds into Variational Autoencoders (VAEs). Classifier-based decision systems utilizing optimization methods may implicitly represent feature relationships to capture information such as shortest distances. However, these systems often operate based on correlation rather than prediction accuracy, potentially learning spurious correlations and yielding erroneous outcomes (Caruana et al. 2015; Joshi et al. 2019). [This issue arises due to the failure to encode causal relationships between variables, which has exposed] the importance of learning and deploying causal models in practical decision-making systems. In a causal decision-making system, one of the primary goals of causal decision systems is to evaluate outcomes under different "treatments" and use interventions corresponding to the treatment that improve the outcome (Joshi et al. 2019, p5).

As a concrete implementation method, (Joshi et al. 2019) explicitly utilizes a causal model employing confounders rather than relying on counterfactual-based Structural Causal Models (SCM) (Karimi, Barthe, et al. 2020; Karimi et al. 2021, 2023; Kügelgen et al. 2023) to learn manifolds. In other words, [the causal model] in developing such decision-making systems attempts to (approximately) learn in the presence of hidden confounders (Louizos et al. 2017; Madras et al. 2019) by estimating these confounders (Joshi et al. 2019, p5). During the learning process, the back-door adjustment formula is employed (Pearl et al. 2016). The main assumption made [here] is that hidden confounders can be reasonably estimated via latent variable models leveraging (approximate) learning algorithms (Joshi et al. 2019, p5).

Meanwhile, (Joshi et al. 2019) highlights the limitations of methods that approximately estimate hidden confounders as follows:

"While methods that approximately estimate hidden confounders are an empirical improvement over classification systems (Louizos et al. 2017), a myriad of issues ranging from mis-specification of the underlying causal model, approximations used for tractability of the latent variable estimation, and selection bias in the data can cause causal models to be less than perfect. Also, while more accurate, outcomes can still be undesirable for many individuals scrutinized under such systems (providing treatment still does not improve outcome even though on average, treatments are effective)" (Joshi et al. 2019, p6).

### B. (Tosi et al. 2014) and (Arvanitidis et al. 2018)

On the other hand, the issues inherent in the optimization-based distance calculation method of (Joshi et al. 2019)—namely, dimensional expansion being performed through submanifold settings while feature relationships are separated and processed as a virtual surface along the external boundaries of the submanifold—can be linked to the identifiability problem in latent variable models, where different representations can give rise to identical densities (Arvanitidis et al. 2018; Bishop 2006). In a mathematical and technical sense, such virtual surface can be understood as straight lines in the latent space Z of VAEs or Generative Adversarial Networks (GANs). [However, because the] straight lines in Z are not shortest paths in any meaningful sense and therefore do not constitute natural interpolants (Arvanitidis et al. 2020, p1) (Figure 1).

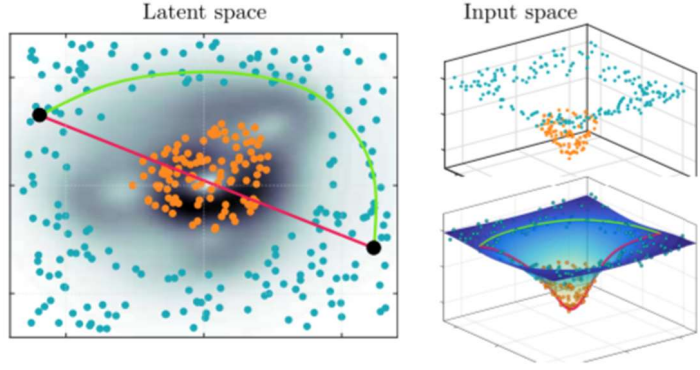

Figure 2. Shortest paths on the surface spanned by the generator do not correspond to straight lines in the latent space, as is assumed by the Euclidean metric. The image is adapted from (Arvanitidis et al. 2018).

To address this, (Joshi et al. 2019) reformulated the aforementioned mathematical and technical challenges of optimization into a logical and technical problem of causality. [Specifically, they noted that] classification based decision-making systems are limited in that they do not encode causal relationships between variables while potentially learning spurious correlations (Caruana et al. 2015, p5). Through latent variable models capable of utilizing approximate learning algorithms, they attempted to resolve the issue of causality by estimating hidden confounders. Nevertheless, this approach addresses mathematical and technical challenges indirectly, by reframing them as logical and technical problems. Therefore, there remains a need to solve these challenges directly as mathematical and technical issues.

In this regard, it has been proposed to endow the latent space with a Riemannian metric such that curve lengths are measured in the ambient observation space $\chi$ (Arvanitidis et al. 2018, 2020; Tosi et al. 2014). This approach immediately solves the identifiability problem. In other words, this ensures that any smooth invertible transformation of $Z$ does not change the distance between a pair of points, as long as the ambient path in $\chi$ remains the same (Arvanitidis et al. 2020, p1).

This can be expressed in mathematical terms as follows,

$$\text{Length}[f(\gamma_t)] = \int_0^1 \|\dot{f}(\gamma_t)\| dt = \int_0^1 \|J_{\gamma_t} \dot{\gamma}_t\| dt, \quad J_{\gamma_t} = \frac{\partial f}{\partial z}\bigg|_{z=\gamma_t,} \quad (2)$$

$$\|J_{\gamma_t} \dot{\gamma}_t\| = \sqrt{(J_\gamma \dot{\gamma})^T (J_\gamma \dot{\gamma})} = \sqrt{\dot{\gamma}^T (J_\gamma^T J_\gamma) \dot{\gamma}} = \sqrt{\dot{\gamma}^T M_\gamma \dot{\gamma}}, \quad (3)$$

Here, $J_\gamma \dot{\gamma}$ is the Jacobian matrix of $\dot{f}(\gamma_t)$) Furthermore, in Eq. (2), $M_\gamma = J_\gamma^T J_\gamma$ is a symmetric positive definite matrix that defines the Riemannian metric.

Furthermore, in the context of Riemannian geometry, the Jacobian matrix $J_\gamma$ is always positioned on the left and right of the Riemannian metric, in the form of a transposed matrix and an original matrix, respectively. Additionally, the Riemannian metric placed between them represents a level that is one higher than the corresponding dimension. This implies that the geometric structure of this Riemannian metric, when represented in terms of the coordinate plane, should have the form of a point, which symbolizes a virtual 'linear (higher dimension)', enabling a sort of embedded form to contain 'lower dimensions' (Kim and Shim 2025, p21). This embedded form implies that the optimization process in this framework does not encounter the problem identified in (Joshi et al. 2019), where dimensional expansion (reduction in the coordinate plane) and feature relationships are separated and therefore, the former is accepted as a submanifold; and the latter is processed as a virtual surface. Instead, through the process of embedding, optimization is performed within the intrinsic space of the expanding dimensions (manifold), not along the external boundaries.

In summary, endowing a Riemannian metric with the latent space and utilizing it enables the term $\|J_\gamma \dot\gamma\|$ in Eq. (2) to be locally expressed as a single point even in the ambient (input) space as well as the latent space. Thus, it becomes possible to measure the curve length in the ambient space $\chi$ by leveraging the overlapping properties between the ambient (input) space and the latent space.

## C. (Arvanitidis et al. 2020, 2021)

The method proposed by (Arvanitidis et al. 2018)—endowing the latent space with Riemannian metric, by means of the Jacobian and measuring it in the ambient (input) space *via* Euclidean measure—raises an important question. As seen in Eq. (2), the term of $\|J_\gamma \dot\gamma\|$ is locally just a single point in the ambient (input) space. Thus, for Euclidean measure to apply, it requires a process to normalize it to unit length, which is represented in Eq. (1) as an integral between 0 and 1. This process is sufficient for measuring curve lengths in the ambient (input) space $\chi$. However, extending this to stochastic cases introduces complications. In the stochastic case, the distance measured in the ambient (input) space needs to be brought back to the latent space (requiring a pull-back process), but there is no clear explanation for this. Measuring distances in the ambient (input) space and bringing the results back to the latent space are fundamentally different tasks.

To address this issue, the subsequent work of (Arvanitidis et al. 2020) defines the inverse function of the Jacobian and uses it to map the metrics and their measurements from the ambient (input) space into the latent space. Namely, (Arvanitidis et al. 2020) entitles the ambient (input) space with Riemannian metric, enabling the use of pull-back metrics to measure curve lengths in the latent space. Through this process, distortions and curvatures of the ambient (input) space can be reflected in the measurement of curve lengths in the latent space. In other words, because (Arvanitidis et al. 2020) endows the ambient (input) space with a Riemannian metric, the framework allows the learning of a Riemannian metric (referred to as the ambient metric) based on the ambient (input) space, enabling the measurement of data density and curvature. This approach leverages high-dimensional data information in the ambient (input) space.

However, (Arvanitidis et al. 2020) also has several critical limitations. First, modeling precision using a Radial Basis Function (RBF) is challenged by the difficulties in selecting parameters and determining the number of components $K$. Second, there may happen the metric instability in the latent space due to the curse of dimensionality. Third and most importantly, the computational cost of calculating the Jacobian and its derivatives is a major concern. Therefore, (Arvanitidis et al. 2021) proposes a novel locally conformally flat Riemannian metric as a simpler, more efficient, and robust alternative to the pull-back metric. Such a novel metric, since it learns based on prior distributions by making use of an Energy-Based Model (EBM) enabling to grasp the structure of the whole data simply *via* the process of allocating energy in the latent space, eliminates the need to use RBFs separately in the ambient (input) space to measure data density and incorporate higher information, as in (Arvanitidis et al. 2020).

### D. (Dominguez-Olmedo et al. 2023)

(Dominguez-Olmedo et al. 2023) modifies the geometry used for calculating distances to be suitable for leveraging submanifolds. In terms of technical perspective, it is similar to (Joshi et al. 2019). Mathematically, however, it aligns with the manifold hypothesis employed by (Arvanitidis et al. 2018, 2020, 2021; Tosi et al. 2014), which uses Riemannian metrics and their pull-back variants. What distinguishes (Dominguez-Olmedo et al. 2023) is its logical foundation: Characteriz[ing] the data manifolds entailed by Structural Causal Models (SCMs) that, besides modeling the data distribution, incorporate additionally knowledge about the causal relationships between the variables of the modeled system (Dominguez-Olmedo et al. 2023, p1).

In other words, while both (Joshi et al. 2019) and (Dominguez-Olmedo et al. 2023) attempt to address mathematical and technical challenges by means of optimization and Riemannian manifolds, they also try to solve the problems left in mathematical and technical challenges by reframing the abovementioned problems as logical and technical challenges — a causation.

### E. (Dombrowski et al. 2024) and (Pegios et al. 2024)

(Dombrowski et al. 2024) and (Pegios et al. 2024) propose a novel framework for generating counterfactual explanations.

First, (Dombrowski et al. 2024) builds on the loss optimization approach proposed by (Wachter, Mittelstadt, and Russell 2017) by improving the optimization process in the input space. They achieve this by generating the latent space through normalizing flow, which enables coordinate transformation (i.e., mapping to the latent space) by means of one-to-one diffeomorphisms. In the process, this latent space is trained to align with the data manifold. They then introduce a method to generate counterfactual trajectories that remain on the data manifold by performing gradient ascent optimization (Eq. (4)) in the latent space:

$$x^{(i+1)} = x^{(i)} + \eta \frac{\partial f_t}{\partial x}(x^{(i)}), \quad (4)$$

where $\eta$ is step size, $f_t(x^{(i)})$ is target class function, $x^{(i)}$ is data in input space $\chi$, and $x^{(i+1)}$ is recursive result of gradient ascent (Dombrowski et al. 2024).

$$z' = z - \eta \cdot \nabla l(c(g(z)), y), \quad (5)$$

Both Eq. (4) and Eq. (5) are gradient ascent methods representing optimization processes. However, Eq. (4) is used in the ambient space, while Eq. (5) is applied in the latent space. Where $\eta$ is step size, $z \subset Z = \mathbb{R}^d$ of a generative model moving through latent space, and $c(g(z))$ is classifier with respect to the decoder output (Pegios et al. 2024).

Despite the use of Riemannian metrics, (Dombrowski et al. 2024) approach compresses all data information into the latent space, effectively functioning as a Euclidean metric during the path optimization. This limitation may prevent the model from fully capturing the non-linearity of the latent space or the complex geometric structures of the data manifold. As a result, the issues such as vanishing gradients near decision boundaries can arise. Counterfactual trajectories might deviate from the data manifold, causing extrapolation beyond the range of data and resulting in unrealistic counterfactual paths (Pegios et al. 2024).

To address these limitations, (Pegios et al. 2024) builds upon the Riemannian geometry tools introduced by (Arvanitidis et al. 2018, 2021). They generate counterfactual trajectories in the latent space that respect the geometry of the data manifold by employing pull-back Riemannian metrics that consider the interaction between the ambient and latent spaces. In other words, (Dombrowski et al. 2024) optimize by the Eq. (4) based on Eq. (1) of (Wachter, Mittelstadt, and Russell 2017) by means of Stochastic Gradient Descent (SGD) approach in the latent space Z; while (Pegios et al. 2024) operate the optimization process that reflects both the data manifold and the classifier's complex decision boundaries by the following Eq. (6), using Riemannian Stochastic Gradient Descent (RSGD) approach. This optimization is performed as follows.

$$z' = z - \eta \cdot \frac{r}{|r|_2}, \quad (6)$$

where $r = M_{\mathbb{Z}}^{-1}(z)\nabla_z f_y(z)$, reflecting the latent code to all surface and $f_y(z) = l(c(\mathbb{E}[g_\epsilon(z)]), y)$ is classification spanning latent space $Z$.

Furthermore, (Pegios et al. 2024) distinguishes between basic pull-back metrics and enhanced pull-back metrics in using RSGD approach. Specifically, the basic pull-back metric performs counterfactual explanation optimization process, using RSGD in Eq. (6) based on the pull-back metric Eq. (7) derived from the methods of (Arvanitidis et al. 2018). In contrast, the enhanced pull-back metric utilizes a Sigmoid function Eq. (8) to capture high-dimensional information in the ambient (input) space Eq. (9) and subsequently incorporates this information into pull-back metrics (Eq. (10) and (11)) to optimize counterfactual explanations process, using RSGD-C.

$$M_{\mathbb{Z}}(z) = \mathbb{E}_\epsilon[J_{g_\epsilon}(z)^T J_{g_\epsilon}(z)] = J_\mu^T(z)J_\mu(z) + J_\sigma(z)^T J_\sigma(z), \tag{7}$$

where mean value $\mu: \mathbb{Z} \to \chi$ and standard deviation $\sigma: \mathbb{Z} \to \mathbb{R}_{>0}$ are parameters of neural networks.

$$c(x) = \text{sigmoid}(w^T h(x)), \tag{8}$$

where $h: \chi \to H = \mathbb{R}^H$ with typically $H \gg D$, and $w \in \mathbb{R}^H$ the weights of the last layer. The mapping $h(\cdot)$ is the learned representation of the final layer, spanning a $D$-dimensional submanifold of $\mathbb{R}^H$ under mild conditions. By endowing $H$, where the classification bounds.

$$M_\chi(x) = J_h(x)^T J_x(x), \tag{9}$$

$$M_{\mathbb{Z},\chi}^\epsilon(z) = J_{g_\epsilon}(z)^T M_\chi(g_\epsilon(z)) J_{g_\epsilon}(z), \tag{10}$$

$$\widehat{M}_{\mathbb{Z}}(z) = \mathbb{E}_\epsilon[M_{\mathbb{Z},\chi}^\epsilon(z)] \approx J_\mu^T(z) M_\chi(\mu(z)) J_\mu(z) + J_\sigma(z)^T M_\chi(\mu(z)) J_\sigma(z), \tag{11}$$

## 2. Methodological Limitations of Existing Literature on Feature Relationships

### 1) Summary of Discussions – Externalization vs. Internalization and Mathematical-Technological vs. Causal Methodologies

This study has so far reviewed how the relationships among data features are addressed in AI literature, particularly those related to counterfactuals. In so doing, considering that performing counterfactuals is based on the classifier's characteristics, this study categorized the literature into externalization and internalization methodologies, when it comes to grasping the relationships among data features, according to whether they analyze feature relationships by flattening them into a single dimension relative to the classifier or by treating them based on submanifolds. Internalization methodologies, in particular, face the challenge of dealing with black-box classifiers, which make it difficult to calculate the "distance to the decision boundary" in non-linear spaces (Venkatasubramanian and Alfano 2020). The works of (Arvanitidis et al. 2018, 2021; Dombrowski et al. 2024; Pegios et al. 2024) represent the attempts to solve these issues by employing mathematical methodologies such as optimization and Riemannian metrics (particularly pull-back metrics), as well as technological methodologies like latent variable generative models (e.g., VAEs or GANs) that respect the geometry of data. This study argues that the CFP theory, as a philosophical approach to data science, can provide a new geometric structure for data, when it comes to the attempt to solve the problem of black box classifiers, thereby offering a clearer understanding of the data generation process (Kim and Shim 2025)[2].

However, the challenges posed by black-box classifiers do not necessarily need to be addressed solely through mathematical and technological perspectives. The process of performing counterfactuals inherently involves a causal mechanism in which specific feature values should be altered as causes to produce the outcomes that pass the decision boundary. By considering such causal mechanisms—an inherently logical perspective—it is possible to address the challenges of black-box classifiers. [In other words, one way] to account for the dependencies between features to model the outcome of interest using features that directly figure into the causal mechanism (Barocas et al. 2020, p83). This is because simply relying on statistically significant features is insufficient to identify causal relationships when predicting or modeling desired outcomes (e.g., credit approval) (Caruana et al. 2015). Instead, the model must include features that causally influence the outcome. To solve this problem through causal mechanisms, however, it is necessary to "employ" causal mechanisms into features, and for this to happen, all such constraints can be identified in advance, which assumes that the actions necessary to change specific feature values will always be self-evident (Barocas et al. 2020). For example, (Joshi et al. 2019) attempts to incorporate causal mechanisms into features by approximating learning in the presence of hidden confounders, while (Dominguez-Olmedo et al. 2023) characterizes data manifolds by embedding SCMs.

This study proposes addressing the problem of black-box classifiers through causal methodologies by introducing the CFP theory. The study asserts that the common cause problem[3]—often considered as a conundrum of application—can be better understood by extending beyond the limited spatial or temporal coordinates of Euclidean space (Buckner 2020) into Hilbert space and utilizing backward causation in the latter space. By doing so, it argues that emergent features are closely associated with this approach and introduces the CFP theory as a theoretical and methodological framework to address this issue.

---

[2] For further details, refer to Section IV. Application of CFP Theory and (Kim and Shim 2025).

[3] The principle of common cause assumes that when there is a correlation between specific events, a common cause exists to explain this correlation. For instance, if two independent events show a correlation, it is generally attributed to a shared common cause. In classical limited spatial or temporal coordinates, such common causes are assumed to always lie in the past, as they rely on the asymmetry of causation and the temporal precedence of events. However, if the possibility of backward causation is accepted, common causes could also exist in the future or in non-spatiotemporal domains (Price 1997; H. Reichenbach 1956). Further discussion on this topic is provided later in this study.

## 2) Limitations of Causal Methodologies in Internalization Models – The Issue of "Common Cause"

Both (Dominguez-Olmedo et al. 2023) and (Joshi et al. 2019) adopt different approaches to calculating distances using manifolds—(Joshi et al. 2019) approximates hidden confounders, while (Dominguez-Olmedo et al. 2023) employs SCMs. Despite these differences, both approaches share the goal of closely considering the dependency between features, using causal models. However, a clear limitation exists in their ability to address the issue of common cause.[4] In other words, whether we consider (Joshi et al. 2019)'s method of estimating confounders or the structural assumptions underlying SCMs proposed by (Dominguez-Olmedo et al. 2023; Pearl et al. 2016),[5] the limitations of these approaches are inevitable. For example, the estimation of confounders using the back-door adjustment formula (Pearl et al. 2016) ultimately leads to a matter of choice and must rely on approximations through learning. Similarly, the structural assumptions of SCMs, which combine graphical models with causal logic, imply a sort of strict constraint that expresses one-sided causal logic as definite and intuitive, thereby making it

---

[4] Refer to Section 1. 3) (a) (Ustun et al. 2019). More details in this case are as follows: The issue of common cause is exemplified in a loan case. Changing a feature (e.g., career changes, denoted as $C$) impacts other features differently—positively affecting another feature of salary ($D$) and negatively affecting another one of employment period ($E$). Here, $C$ acts as a common cause influencing both $D$ and $E$. This situation is described as "$C$ screens off $E$ from $D$," represented mathematically as $P(E|D\&C) = P(E|\sim D\&C)$. When a common cause exists, a screening relationship is established; in its absence, $C$ merely mediates the causal relationship between $D$ and $E$ (Reichenbach 1971).

[5] When it comes to the structural assumptions underlying SCMs, SCMs consist of a set of variables $U$, $V$, and functions $f$, where $f$ assigns values to each variable in $V$. The assumptions for $f$ are as follows:

(1) Variables in $U$ are exogenous, external to the model, and their origins are unexplained.

(2) Variables in $V$ are endogenous, and each endogenous variable is a descendant of at least one exogenous variable.

(3) Exogenous variables cannot descend from other variables, particularly endogenous ones. They are root nodes in the graphical representation.

Based on these assumptions, if the values of all exogenous variables are known, $f$ can completely determine the values of all endogenous variables (Pearl et al. 2016). Meanwhile, in terms of treating an action in the real world as an intervention, it is possible to make the assumption that the actions in the real world correspond one-to-one with the interventions on endogenous variables in the SCM. This makes it impossible to account for confounded or correlated interventions and, therefore, to establish common causes (Karimi et al. 2021).

impossible to establish common causes.[6] As such, as long as methods rely on approximating confounders or embedding SCMs into technological tools such as VAEs or GANs, the issue of common cause remains ultimately challenging to resolve.

**3) Overlooked Aspects of Feature Relationships – Newly Emergent Features**

The existing literature's inability to fully capture feature relationships such as "common causes," due to the implicit assumption established in the mathematical, technological, and causal methodology, implies the limitations imposed by predefined conditions inherent in the existing frameworks. These methodologies are bound by the limited spatial or temporal coordinates, such as those of Euclidean space (Buckner 2020). As a result, causal constraints and features must be self-evident and fully identifiable in advance (Barocas et al. 2020). Consequently, conventional causal mechanisms only consider the causes of an outcome retrospectively, once the outcome has been determined. However, this retrospective perspective — that is to say, the case which should be determined based on a certain time — may appear unreasonable in the dynamic processes where features continuously interact with one another, even altering the outcomes, and even the order of cause and effect, as events unfold.

From this perspective, "common cause" can be understood as a problem arising from interpreting the features, overlooking the relational dependency between the features generating dynamic configurations, solely through a retrospective lens and within the constraints of limited spatial-temporal coordinates. In other words, the notion of "common cause" can miss the complex interactions between multi-dimensional relationships due to its reliance on simplistic reductionism under the plane-conditional environments such as two-dimensional spatial-temporal models.

To better understand this, let us more specifically analyze the loan case that addresses the issue of common cause. Assume an individual applying for a loan has the features such as career changes, salary, bank account balance, and employment period. Also, suppose the loan applicant needs to increase their salary by $5,000 or maintain a bank account balance exceeding $10,000 to secure the loan. If the applicant receives a $3,000 bonus through a career change but does not reach the $5,000 salary increase threshold, they will not qualify for the loan. However, if the $3,000 bonus is deposited into their bank account, surpassing the $10,000 threshold, the bonus significantly influences another cause of loan approval. Here, the career change acts as the common cause (feature $C$), salary increase as its effect (feature $D$), and bank account balance as another effect (feature $E$). Therefore, salary (feature $D$) and bank account balance (feature $E$) exhibit a correlation, not causation.

---

[6] Graphical models are often combined with SCMs because our knowledge of causal relationships tends to be qualitative, as represented by graphical models, rather than quantitative. For instance, we know that gender influences height, and that height influences basketball performance, but we hesitate to assign numerical values to these relationships. Graphical models enable the intuitive understanding of causality in such cases, complementing SCMs (Pearl et al. 2016).

Additionally, consider a scenario where, unexpectedly, the applicant lends $3,000 to a sibling who has caused a car accident, and cannot deposit the bonus into their bank account. The sibling repays the money a week after the loan assessment. While it is legally valid to deny the loan based on the situation at the assessment time, the justification for the decision may remain susceptible. This case exemplifies how unforeseen, temporary events can disrupt the static conditions assumed in conventional models.

What is critical here is that the common cause (feature $C$) does not remain fixed as a single past event but can extend into "future possibilities" through the changing conditions and interactions among features. For instance, in the loan case, while a career change aimed at salary increases may lead to a higher bank account balance, a lottery win resulting in a higher bank balance (feature $E$) could prompt the applicant to consider career changes (common cause $C$) for a more relaxed lifestyle. In such scenarios, the temporal directionality of causation between career changes ($C$) and bank account balance ($E$) may reverse, with salary ($D$) decreasing. Conversely, if the lottery win inspires the applicant to save more for future business opportunities, they may seek a career change ($C$) to increase their salary ($D$). These examples illustrate how the mutual influence between career change ($C$) and bank account balance ($E$) can vary depending on the circumstances of the loan applicant, which in turn can lead to either an increase or a decrease in another feature—salary ($D$). Furthermore, rather than understanding this solely as a reversal in the directional relationship between career changes (common cause $C$) and bank account balance (result $E$) that leads to a change in salary (result $D$), it can also be interpreted that the increase or decrease in salary (result $D$) and the growth of the bank account balance (result $E$) generate the future possibility of career changes (common cause $C$) as an emergent feature.

This temporal relativity and dynamic nature of feature interactions are further evident in AI models for autonomous vehicles. For example, consider an autonomous vehicle navigating a highway to avoid a pedestrian. The cluster of features $C$ could represent the car's state (e.g., speed, position, braking power), the cluster of features $E$ the pedestrian's state (e.g., position, speed, behavior), and feature $D$ the vehicle's perception of collision risk. In a conventional causal framework, the cluster of features $C$ might explain a sudden car's reduction in speed, while the cluster of features $E$ account for the pedestrian avoiding the car. However, unlike conventional vehicles, autonomous vehicles leverage collision risk (feature $D$) to detect pedestrians, calculate collision probabilities, and decide to reduce speed. Feature $D$ thus influences both the cluster of features $C$ and $E$, establishing a correlation between the latter two.

Moreover, the car's state and the pedestrian's behavior continuously interact, producing new outcomes. Instead of fixing feature $D$ (collision risk) as a single past event, it can be understood as an emergent feature generated through the interactions between the cluster of features $C$ (the car's state) and $E$ (the pedestrian's behavior). For example, the autonomous vehicle may reduce speed (the cluster of features $C$) upon detecting the pedestrian's movement (the cluster of features $E$), reversely prompting the pedestrian to adjust their behavior (the cluster of features $E$) in response. Within these interactions, collision risk (feature $D$) dynamically adjusts in real time. In this sense, feature $D$ can be considered both a "common cause" and a newly emergent feature representing future possibilities.

# 3. Shift in Discussion

## – Common Cause as a Limitation of Existing Approach and New Emergent Feature for Overcoming Limitation

### 1) Need for New Conditions and Causal Logic

The interpretation of common cause-related features into emergent features, as discussed above, may be contested as disregarding the temporal precondition inherent in mathematical and technological methodologies, which are fundamentally bound by limited spatial or temporal coordinates (Buckner 2020). Furthermore, assuming the generation of new features contradicts the fundamental assumption that all causal constraints and features must be fully identifiable in advance (Barocas et al. 2020). In other words, within a deterministic framework, where all conditions are predefined and identifiable, the relationship between features—particularly causation—should at least be judged under the premise of (probabilistic) independence among features.

Of course, the validity of the fundamental assumptions underlying the mathematical, technological, and causal methodologies is an overarching premise for almost all agendas related to AI technology. However, an easily overlooked yet crucial fact should be acknowledged. One of the most critical characteristics inherent in AI technology is the issue of explainability, closely related to a black-box phenomenon. More specifically, the emergence of a new perspective of "explanation vs. prediction" suggests that, in some cases, fundamentally new theoretical attempts that break away from the assumptions embedded in conventional mathematical, technological, and causal methodologies are required for ultimate resolution. For example, as previously discussed, if we adhere strictly to a deterministic worldview and attempt to analyze feature relationships solely within a limited Euclidean spatial-temporal coordinate system, we inevitably fail to capture newly generated emergent features. Consequently, our understanding of feature relationships remains incomplete. Similarly, if we adopt an approach like that of (Dominguez-Olmedo et al. 2023) or (Joshi et al. 2019), where feature relationships are addressed by expanding dimensions quantitatively within a limited Euclidean spatial-temporal coordinate system, we may approximate the topology related to feature relationships. However, this approach still fails to capture the aforementioned newly emerging features. The failure to detect emergent features due to the mathematical limitations of Euclidean spatial-temporal models should not be viewed as the fallacy of mathematical tools but rather as a gap in the mathematical tools currently in use—one that must be identified and overcome.

Thus, if common causation-related features can be understood as emergent features within a dynamic framework where feature relationships continuously change over time, then the current planar conditions—such as those found in two-dimensional spatial-temporal models—are insufficient. A new condition capable of representing complex interactions in a multi-dimensional context is required. Moreover, within this new condition, a different causal logic—one that transcends conventional causal

reasoning—should also be adopted.

## 2) Hilbert Space and Backward Causation

   To address this issue, this study proposes the adoption of Hilbert space instead of Euclidean space, and the incorporation of backward causation in place of the strict asymmetry imposed by conventional causal reasoning within limited spatial-temporal coordinates. Under a Euclidean spatial-temporal framework, a common cause (e.g., $D$, as seen in the autonomous vehicle case discussed earlier) temporally precedes its effects ($C$ and $E$), following a directional relationship: $D \rightarrow C$, $D \rightarrow E$. However, by adopting Hilbert space and backward causation, this study suggests that the effects ($C$ and $E$) interact to dynamically generate a new common cause ($D$). Within such a framework, the already done past outcomes ($C$ and $E$) can be connected and interacted with each other and linked into an upward single dimension, the following likely future possibility ($D$).

   Hilbert space is a complete set of vector space with an inner product, which can be defined not only in finite dimensions but also in infinite dimensions (Conway 2019; Rudin 2012). This property allows it to reflect the high-dimensional nature of data and to handle non-Euclidean properties. Traditionally, Euclidean space primarily uses diagonal matrices to represent data dimensions. However, it is important to note that the interactions beyond the diagonal components may be challenging to express accurately within Euclidean space. In other words, non-diagonal components in a matrix can represent the interactions that are difficult to capture within Euclidean space. Therefore, an alternative approach in Euclidean space is to approximate these components by replacing them with diagonal elements. However, it remains questionable whether this substitution accurately represents the underlying interactions. Unlike these constraints, Hilbert space—although still subject to the constraints of linear matrix operations and the convergence conditions of Cauchy sequences—provides a richer representation of interactions among data components beyond the diagonal elements of a matrix (Kim and Shim 2025).

   Given this foundation, the concept of backward causation can be incorporated naturally into Hilbert space. Backward causation posits that effects can influence their causes, fundamentally reversing the conventional time-ordering of causality (Rovelli 2018). To understand this concept more precisely, it is essential to examine the notion of causation asymmetry. Conventional causal reasoning assumes that causation is asymmetric, meaning that causes precede their effects and push them "forward" in time. While conventional views often equate causality with temporal asymmetry, these two concepts must be distinguished (Dummett 1964; Price 2007). In other words, even if causation asymmetry holds, backward causation can still be valid, and Hilbert space provides the mathematical foundation that makes this possible.

   The fact that the features understood as common causes can be reinterpreted as emergent features through the application of backward causation within Hilbert space is supported by quantum entanglement phenomena in quantum mechanics. In quantum states, the position and momentum of a particle depend on observation, leading to an indistinct boundary between past and future.[7] Thus, if the outcome features $C$ and $E$ in the autonomous vehicle case are considered as the measurement results of two entangled particles, and the common cause feature $D$ is understood as a quantum entangled state, then from the classical perspective of causation, the entangled state ($D$) is assumed to have been generated in the past, thereby a posteriori determining the outcomes $C$ and $E$. However, under a Hilbert space framework, even though $C$ and $E$ have already occurred as results, their interactions can dynamically generate $D$ (the entangled common cause) as a future possibility, which will form an upward dimension consisting of $C$ and $E$ as its respective components. In other words, within Hilbert space, $D$, $C$, and $E$ are not temporally separated but exist simultaneously as an upward dimension constructed through probabilistic operations and interactions among state vectors.

### 3) Theoretical Foundation for Implementing a Causal Model in Accordance with Hilbert Space – CFP Theory

While interpreting the problem of common cause within Hilbert space through backward causation enables an understanding of emergent features—further supported by quantum entanglement—there remains an open question: How can backward causation be concretely implemented into a causal model within Hilbert space? Addressing this issue requires a mathematical and technological approach, such as utilizing the state function of quantum-mechanical Hilbert space (e.g., the Schrödinger equation) or integrating Hilbert space with general relativity to relativize the directionality of time or combining quantum-mechanical approaches (Kastner 2012; Rovelli 2018; Wharton 2007). However, beyond these approaches, the foremost priority is to establish a theoretical foundation for implementing a causal model that aligns with the properties of Hilbert space.

To establish this philosophical and theoretical foundation, this study introduces CFP theory, which formulates a new relational concept known as "expansion of dimensions accompanied by qualitative transformations." This concept is further developed through two conceptual hypotheses, providing a structured theoretical framework for understanding emergent features.

To achieve this, this study takes a fundamental perspective on the understanding of AI. Since AI research is closely related to statistical modeling in data science, it first briefly examines the two cultures of statistical modeling proposed by (Breiman 2001): the data modeling culture and the algorithmic modeling culture. Following this, to clarify the distinctions between data science and AI, specifically speaking, Deep Neural Networks (DNNs), this study defines the concept of "data-algorithm relationship modeling" as an inherent statistical modeling approach of DNNs. Based on this, it conceptualizes the developmental phase of DNNs—characterized as an expansion of dimensions accompanied by qualitative transformations—as a new paradigm for understanding feature relationships in AI systems.

---

[7] This is understood as non-locality in classical space-time framework.

## III. Modeling in Data Statistics and Unique Modeling of DNNs

### 1. Traditional Statistical Modeling in Data Statistics — Data Modeling Culture and Algorithmic Modeling Culture

**1) Comparison Between Data Statistics and DNNs**

As previously mentioned, artificial intelligence research is fundamentally based on data statistics. Regarding the distinction between the two, it has been stated that "artificial neural networks are a subfield of machine learning, which is in turn a subfield of both statistics and artificial intelligence" (Skansi 2018, preface v), and also that "machine learning is a marriage of statistics and knowledge representation" (Flach 2012, preface xv).

(Srećković et al. 2022) discusses the similarities and differences between data statistics and machine learning as follows:

> What needs to be kept in mind is that there is no sharp division between traditional statistics and machine learning, and that there is a large gray area. There is no single criterion that can clearly separate ML methods from statistical methods, and the very difference between the two is a matter of debate (Bzdok 2017; Bzdok et al. 2018; Bzdok and Yeo 2017). There are many methods and tools (e.g. linear regression, bootstrap…) that are used in both of these traditions of modeling. Even though the area of overlap between the two disciplines is large, some general characteristics are quite different. For us, the most relevant difference is the way of using traditional statistics and the way of using machine learning. For traditional statistics, standard models rely on the representation of underlying causal mechanisms, and they are used for retrospective testing of an already existing set of causal hypotheses. On the other hand, ML models are constructed based on data instead of theoretical assumptions about the target system. The purpose of such models is primarily forward-looking, i.e. to predict new observations (Shmueli 2010). In short, while traditional statistical models are used to detect causal information of explanatory value, ML models are used to detect associative information of predictive value (Srećković et al. 2022, p165-166).

Understanding these differences between DNNs and data statistics—specifically, the phenomenon where DNNs employ more complex and unstructured data than traditional statistical methods, focus on prediction (rather than explanation), and achieve superior performance but lack interpretability—requires further exploration. This issue ultimately leads to the challenge of balancing prediction and explanation in the use of AI, particularly in DNNs.

A key concern arises when the predictive capabilities of DNNs far exceed both the quantitative and qualitative capacity of human prediction and yet humans cannot fully comprehend these predictions. This could instill a sense of fear that AI is controlling or dominating human decision-making processes to people. This concern is also one of the reasons why people are wary of the emergence of Artificial General Intelligence (AGI). Therefore, a fundamental understanding of modeling methods in data statistics is a prerequisite for addressing this issue. Since data statistics serve as the foundation of DNNs, developing a deeper understanding of DNNs' unique modeling approach is crucial to structuring and making sense of the distinct phenomena that set DNNs apart from traditional data statistics.

## 2) Data Modeling Culture and Algorithmic Modeling Culture

Regarding modeling methods in data statistics, since (Neyman 1939, p55) mention that [applying statistical concepts to data requires] some system of conceptions and hypotheses, the consequences of which are approximately similar to the observable facts, several statisticians have acknowledged a fundamental distinction in terms of the uses of statistics, such as the theory-driven approach vs. data-driven approach (Boge 2023) or data modeling culture vs. algorithmic modeling culture (Breiman 2001) (see, for instance, (Breiman 2001; Davies 2014; D. Hand 2019; D. J. Hand 2009; Lehmann and Lehmann 2012; Shmueli and Koppius 2011). This study adopts (Breiman 2001) distinction as the foundation for understanding modeling methods in data statistics.

[(Breiman 2001, p99) defines] two goals in analyzing the data: [one is] Prediction. To be able to predict what the responses are going to be to future input variables; [the other] is Information. To extract some information about how nature is associating the response variables to the input variables. [In sequence,] there are two different approaches toward these goals.

[First, the data modeling culture] assumes a stochastic data model for the black box. For example, a common data model is that data are generated by independent draws from response variables = $f$ (predictor variables, random noise, parameters). The value of the parameters is estimated from data and the model then used for information and/or prediction. [Model validation is performed through] goodness-of-fit tests and residual examination. [In contrast, the algorithmic modeling culture] considers the inside of the box complex and unknown, treating the data-generating mechanism as inherently unknown. Their approach is to find a function $f(x)$—an algorithm that operates on $x$ to predict the responses $y$. (Breiman 2001, p199). In other words, instead of attempting to model this mechanism explicitly, the analysis in this culture focuses on finding an algorithm that operates on data $x$ as a sort of appropriate pattern such as decision trees or neural nets (model validation is based on predictive accuracy). The algorithmic modeling culture has rapidly advanced, particularly in the fields such as machine learning.

The data modeling culture has the advantage of providing clear information about nature by assuming a stochastic model within the black box. This approach ensures predictive reliability. However, a major drawback is that simplifying complex natural mechanisms can lead to incorrect conclusions. In particular, for complex and unstructured datasets, the prior assumptions of traditional modeling methods can result in misleading conclusions about underlying mechanisms—conclusions that cannot be validated solely through formalized model fit tests (Breiman 2001; Srećković et al. 2022). Conversely, the algorithmic modeling culture offers the merit of higher predictive accuracy for complex datasets. However, since it focuses on appropriate patterns for producing the desired outcomes rather than constructing explicit data models, it inherently faces limitations in ensuring predictive reliability.

## 3) Summary

Both the data modeling culture and the algorithmic modeling culture have their respective significance in drawing conclusions from data. In particular, since the algorithmic modeling culture focuses on finding appropriate patterns related to data for producing the desired outcome and has rapidly developed in the fields such as machine learning, it is more closely aligned with the structure of DNNs. However, one critical aspect overlooked in both statistical modeling approaches in data statistics—whether generating a specific model from data (data statistical modeling) or developing an algorithm for prediction independently of any predefined data model (algorithmic statistical modeling)—is that machine learning, particularly DNNs, undergoes a twofold process (LeCun et al. 2015). Specifically, DNNs first learn an algorithm from training data, and then, based on this, explore an appropriate model through fitting with domain data.

## 4) Appearance of Unique Modeling Approach of DNNs

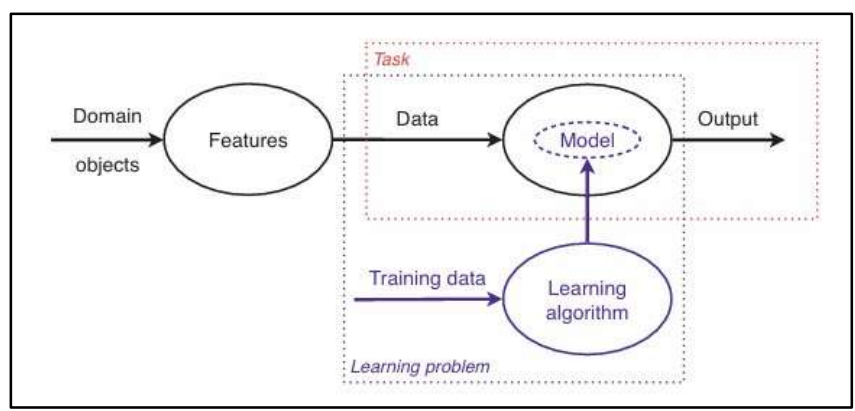

Figure 3. Data-Algorithm Interrelationship Modeling" (Image adapted from (Flach 2012))

Meanwhile, if the main ingredients of DNNs are tasks, models, and features, then DNNs, unlike traditional data statistics, require a clear distinction between tasks and learning problems. As illustrated in Figure 2, the process of modeling, created through the interaction between learning algorithms and data, is not merely a set of parameters (of, for example, a classifier) defined by data features only (Flach 2012). In other words, tasks are addressed by models, whereas learning problems are solved by learning algorithms that produce models (Flach 2012, p11-12).

Despite this, many AI researchers tend to conflate how learning algorithms are developed within the learning problem domain (indicated by the purple dashed area in Figure 2) with how models are developed within the task domain (indicated by the red dashed area in Figure 2). If judged based on linear regression, both processes may appear to follow the format of fitting. However, due to the twofold nature of the process, the contents produced through each fitting procedure inevitably differ. This discrepancy arises because an algorithm trained on training data later combines with domain data, producing different results. As mentioned earlier, Figure 2 illustrates that a model is neither solely defined by the features of domain data nor merely a collection of parameters specified by a learning algorithm.

Taking this one step further, the "model" section in Figure 1, which represents the interaction between domain data and learning algorithms, becomes increasingly complex as the number of nodes in DNNs rapidly expands. Within this increasingly complex interaction, the activation of hidden units emerges (Bau et al. 2018; Boge 2023; Woodward 2005). Unlike linear regression, which measures only linear changes in an algorithm in a directional and retrospective manner, the expansion of nodes based on the twofold process can produce unexpected transformations, akin to synergy effects. Therefore, as DNNs prioritize prediction for improved performance, they inevitably exhibit black-box phenomena, where explanations become increasingly difficult.

Considering the structural nature of DNNs, modeling DNNs should not be understood through the two conflicting cultures of data modeling and algorithmic modeling (Breiman 2001), as in data statistics. Instead, it is more appropriate to focus on the relationship between data and algorithms, framing it as "Data-Algorithm Relationship Modeling." Thus, in the following section, this study defines the concept of "data-algorithm relationship" as the unique modeling approach of DNNs and theoretically examines how this unique modeling process of DNNs leads to a new developmental phase. This theoretical examination will, as previously mentioned, inevitably establish itself as a new theoretical framework that departs from the conventional assumptions inherent in the existing mathematical and logical methodologies.

## 2. A New Understanding of DNNs modeling – Data-Algorithm Relationship Modeling

## 1) Starting Point of the Discussion – Gap between Data Science and Euclidean Space-based Mathematics

To approach a new modeling culture beyond the two traditional statistical modeling cultures discussed by (Breiman 2001), we begin with the gap between data science and Euclidean space-based mathematics. Euclidean space-based mathematics operates under the assumption of a closed world, striving for uniqueness and completeness while establishing its own axioms. Consequently, when employing mathematics as a tool, it must be capable of representation based on limited spatial-temporal coordinates. Furthermore, the expansion of dimensions essentially takes on the form of adding a quantitative scaling process based on a two-dimensional coordinate plane.

In contrast, AI as data science, particularly DNNs, is predicated on an open world, implementing performance by expanding dimensions to establish correlations between the facets of human life expressed as data and the desired models corresponding to them—as previously examined within traditional statistical modeling. Therefore, in order to effectively implement DNNs, it is crucial to analyze "the developmental phase of the relationship between data and models."

This developmental phase inherently exhibits cumulative, multi-layered, and progressive complexity. Thus, if one intends to capture this developmental phase using mathematics as a tool, special caution should be exercised when dealing with dimensional expansion. That is, if AI as data science, which operates under the premise of an open world, progresses based on mathematics as a tool, which can be used in the assumption of a closed axiomatic world, then it is required to acknowledge the fundamental differences between the two when it comes to handling dimensional expansion. However, this difference is often intentionally or unconsciously ignored due to the universality of mathematical language as a scientific tool. Moreover, current scientific systems—rooted in reductionism—face clear limitations in capturing the intrinsic relational development that involves dimensional expansion between subjects and objects under the premise of an open world.

## 2) Dimensional Expansion Accompanied by Qualitative Transformations – Two New Conceptual Hypotheses

In summary, the fundamental gap between AI as data science and mathematics becomes particularly evident when analyzing the relationship between data and model within the open world framework assumed by data science. Therefore, to clarify this distinction, this study argues that the development of the relationship between data and algorithms, which can be grasped as AI's performance, should not be understood merely as a scale-up quantitative dimensional expansion only confined to a planar coordinate plane in terms of mathematics. Instead, it should be conceptualized as an expansion into a higher-dimensional structure accompanied by qualitative transformation - figuratively speaking, how a "plane" can be transformed into a "three-dimensional structure" by undergoing qualitative transformation.

This is especially critical in AI applications beyond simple machinery mechanisms,[8] such as generative AI, where the developmental phase of the relationship between data and models—or, more specifically, "dimensional expansion accompanied by qualitative transformation"—should not be overlooked. This developmental phase results in the emergence of high-dimensional structures as organized complexes, at the level of data feature, beyond the lower-dimensional structures of raw data (Capra and Luisi 2014).[9]

This raises two fundamental questions: First, what exactly does it mean in data science that the relational development between data and algorithms does not simply scale up in a quantitative manner but embody a sort of transformation of planes into three-dimensional structures, being accompanied by qualitative transformation? Next, how can mathematics, as a tool that operates within a closed-world framework, accurately represent AI's developmental process, which aims toward an open-world paradigm?

To address these issues, this study approaches the meaning of "dimensional expansion accompanied by qualitative transformation" through two conceptual guiding hypotheses:

1. The concept of "Creating Relative Space-Time in Relationship (crSTR)" – This hypothesis examines how one-dimensional transforms into another higher dimension, analyzing this process within the relational framework. This concept may also be extended and substituted with Another Bigger I Converging Newly in Days (ABICND) as another entity continuously directed toward relationships.

2. The concept of "Duplex Contradictory Paradoxical Stratified structures of Thorough Closure (Solitude) – Eternal Opening (DCPSs of TC-EO)" – This hypothesis explores the driving force behind dimensional expansion—the catalyst that facilitates qualitative transformation leading to higher-dimensional structures. It suggests that this transformation arises through "Thorough Closure" as a necessary precursor to "Eternal Opening."[10]

---

[8] This was also the case with early linear regression analysis. However, since linear regression analysis (LRA) continues to serve as a fundamental structural component in most DNNs today, it should not be overlooked.

[9] It is crucial to avoid interpreting the high-dimensional structure of data as merely an aggregation of its lower-dimensional structures, a misconception often influenced by reductionism. The conceptualization of data within lower-dimensional structures and within high-dimensional structures entails fundamentally different approaches. In this context, the high-dimensional structure of data should be understood as an emergent and novel construct that is fundamentally distinct from its lower-dimensional counterparts.

[10] Personally, I believe that properly designing this aspect could resolve many challenges in AI and open the door to surpassing the limitations of traditional statistical data modeling. Originally, these two concepts are grounded in relational philosophy, which is founded on an understanding of the intersubjectivity of beings. This relational philosophy draws upon the late French existentialist thought of Henri Bergson and Emmanuel Levinas, as well as the classical Confucian doctrine of Zhongyong (中庸, the Doctrine of the Mean). Additionally, it is rooted in a

Due to the scope limitations of this study, a detailed examination of these hypotheses will be reserved for a forthcoming discussion on "Philosophical Basis of the Data-Algorithm Relationship Model." However, to support the understanding of the first conceptual hypothesis, this study will provide a brief overview of the relevant physical and biological foundations. Regarding the second conceptual hypothesis, this study will examine its correspondence with the concept of "strange loops" discussed in Hofstadter's (1999) seminal work, "Gödel, Escher, Bach: An Eternal Golden Braid" (GEB). By doing so, this study aims to demonstrate that the second conceptual hypothesis can serve as an organizing principle that facilitates the transformation from one dimension to a higher dimension.

### 3) First Conceptual Hypothesis – creating relative Space-Time in Relationship (crSTR)[11]

To grasp how a "plane" transitions into a higher-dimensional structure through a qualitative transformation, it is first necessary to design the concept of Creating Relative Space-Time in Relationship (crSTR). In order to understand this, it is essential to first refine the scientific understanding of the concept of time.

From the perspective of quantum mechanics as modern science, time can be understood as inseparable from space, functioning as a field such as discontinuous waves. Therefore, it is sufficiently possible to construct a theory in which time is hypothesized and defined in relation to space[12]. To devise a way to define time in relation to space within a relational framework, let us first consider the process by which hydrogen and oxygen molecules interact to form water. In this scenario, the formation of a relationship between oxygen and hydrogen necessitates a chemical interaction. It is evident that such a process cannot be merely quantified numerically. However, once this interaction occurs, the chemical bonding process becomes oriented within an existential field, acquiring a specific directionality, which is

---

social science framework known as the Convergent Paradigm (CP), which explores a new theory of organizational systems.

The Convergent Paradigm (CP) posits that a society can achieve genuine harmony and prosperity not merely through the calculation of profits and losses among independent entities, but rather through the formation of organic interrelationships among subjects. Such interrelationships create a foundation that enables entities to continuously acquire a new and evolving identity that transcends individual existence.

Due to the scope limitations of this study, a more in-depth exploration of this topic will be addressed in a separate paper.

[11] For a more detailed discussion on this topic, please refer to the author's other paper, "A Novel Approach to Data Generation" (Kim and Shim 2025)."

[12] It should be noted that the fact that time can be defined together with space does not necessarily imply that time should be understood exclusively through spatialization, as in Einstein's interpretation

expressed in the form of energy (Sucher 1998).

To explain this phenomenon in more detail in terms of physical science, we describe it as follows:

When two chemical substances interact and their electron affinities align, a chemical bond is formed, resulting in the creation of a new compound. In physics, this concept of electron affinity can be understood as the eigenfrequency of each substance. When the eigenfrequencies of two substances resonate, the chemical bond occurs instantaneously at the optimal energy state, leading to the formation of a new material. Eigenfrequency, in this context, represents the inherent time-scale of each substance (Bernath 2020).

The newly formed material, which emerges from the synchronization of the unique time-scales of its constituent substances, can be analogized to a Created Relative Space-Time in Relationship (crSTR). Since crSTR inherently seeks to generate a new existence, this phenomenon can be understood as the emergence of a new system (Auffray et al. 2003; Auffray and Nottale 2008).

Encapsulating the above discussion within the axiomatic frameworks of mathematics and physics is challenging, especially considering the constraints of scientific rigidity. However, it remains theoretically feasible to construct a framework based on this hypothesis. As long as such a theoretical construction does not contradict the established mathematical and physical principles, it poses no fundamental issue. Moreover, there is no limitation in utilizing this hypothesis to develop DNNs modeling techniques within the framework of AI's organizational theory. This is because AI itself has been evolving by incorporating methodologies and conceptual frameworks from various disciplines, for example, symbolism AI derived from logic (Domingos 2015) and connectionist AI inspired by neuroscience (Baek et al. 2021).

Accordingly, from a philosophical perspective, additional explanation is required. However, if we summarize this hypothesis, it can be expressed as follows:

$$A \backsim B \implies crSTR \fallingdotseq ABICND$$

When two entities, *A* and *B*, exist within the same dimension and establish a relationship, they create a relative Space-Time in Relationship (crSTR) (①).

Furthermore, the relative Space-Time acts as a field (場) that constructs a higher-dimensional framework, ultimately converging towards a new entity known as "Another Bigger I Converging Newly in Days (ABICND)" (②).

Table 1. New Conceptual hypothesis for dimensional expansion I.

The 'Generative Dynamic Existence Unit on a Relational Basis,' as summarized in Table 1, can serve as a framework for designing the specific manifestation of dimensional expansion in AI modeling. That is, it conceptualizes how the relationship between data and models unfolds within the premise of dimensional expansion in AI, transitioning from a plane to a higher-dimensional structure through a qualitative transformation.

This study proposes that data and an algorithm, which can be represented within the same dimensional coordinate space, function as:

○ Fragments and traces of human life **(represented as data in Figure 1).**

○ and Human-driven objectives applied to the training data **(represented as the learning algorithm generated from training data in Figure 1).**

Within this framework, both data and algorithms can be regarded as entities in their own right. Consequently, the interaction between data and algorithms, and the developmental process that emerges from this interaction, can be understood not merely as fitting data to an algorithm or discovering implicit patterns inherent in data but as a newly extended dimension **(crSTR, represented as the 'model' in Figure 1).**

This process can be interpreted as the reorganization of human life into a structured form that aligns with the purpose encoded in the algorithm, while, in some cases, modifying even the purpose. Ultimately, this structured representation of human life, which gets along with human intent as embodied in the algorithm, progresses, as time goes on, increasingly toward a new existence—referred to as "Another Bigger I Converging Newly in Days (ABICND)"—which is realized through the activation of hidden units continuously generated by the model (López-Rubio 2021).

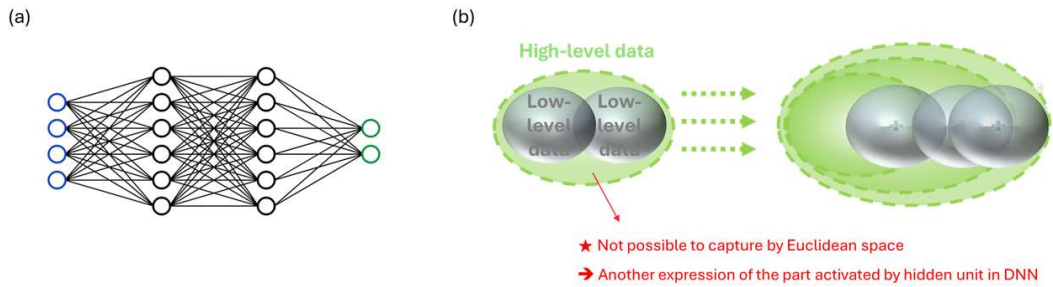

Figure 4. The relationship between nodes connected by complicated lines in (a) and the green part in which crSTRs are manifested in (b) – which represent the common property causing the area of generation. (a) The conventional structure of DNNs. (b) The relationship between low- and high-level data features proceeding from a fixed perspective to an overall perspective (from one scale to another).

**4) The Second Conceptual Hypothesis – DCPSs of TC-EO**

If the first conceptual hypothesis aims to capture the specific manifestation of dimensional expansion accompanied by qualitative transformation, the second conceptual hypothesis seeks to identify the driving force behind such qualitative transformations occurring in AI as data science. As discussed in Section 3, qualitative transformation arises when data and algorithms at the same dimensional level establish relationships, forming a higher-dimensional relative space-time, which in turn enhances performance. In other words, as the crSTR evolves and moves increasingly towards ABICND (Another Bigger I Converging Newly in Days), a hypothesis is required to explain the catalyst that enables qualitative dimensional expansion.

Originally, the DCPSs of TC-EO hypothesis was derived from the Confucian principle of "*Shendu*" (慎獨), a core concept in the Doctrine of the Mean (中庸) in classical East Asian philosophy, particularly Confucianism. However, as previously mentioned in the above Section, a full philosophical explanation of this concept falls beyond the scope of this study.[13] Instead, this study will examine the operational mechanism of the "Strange Loop," as presented in (Hofstadter 1999), to explore how this concept implies the structural framework of DCPSs of TC-EO. By doing so, we can recognize that the development of DCPSs of TC-EO serves as the driving force that allows a system to go beyond its own framework and progress to a higher system. This principle can be further applied to dimensional expansion observed in data-feature relationships, inter-data relationships, and data-algorithm

---

[13] A more in-depth exploration of this topic will be addressed in a separate study, titled "The Philosophical Basis of the Data-Algorithm Relationship Model" (forthcoming).

relationships.

## A. From Gödel's Proof (Typographical Number Theory: TNT)[14] to (TNT+$G_\omega$)–PROOF–PAIR{a, a'}

In the 20th-anniversary edition preface of *Gödel, Escher, Bach (GEB)* (Hofstadter 1999), Hofstadter explores the question, "how is it that animate beings can come out of inanimate matter? (Hofstadter 1999, p.2)". He attempts to answer this by shifting the focus from material composition to abstract patterns, particularly focusing on "the paradox of self-reference".

This paradox of self-reference can be traced back to Russell, and more critically, to Gödel's discovery of "Gödel numbering," which enables the encoding of mathematical statements within numerical systems (Hofstadter 1999). This technique ultimately leads to the formulation of the famous self-referential statement in TNT (Typographical Number Theory):

**"I Cannot Be Proven in Formal System TNT"**[15] **(Hofstadter 1999, p465)**

Hofstadter does not merely interpret this self-referential loop as Gödel's essential incompleteness theorem within formalized mathematics. Instead, he expands the concept by introducing *G* to further reinforce self-reference, allowing for the generalization of proofs.

Specifically, he extends the paradoxical statement to:

**"I Cannot Be Proven in Formal System TNT+*G*" (Hofstadter 1999, p467)**

---

[14] Hofstadter adopted the concept of TNT from the *Principia Mathematica* (Whitehead and Russell 1927), which represents the formalism perspectives of Mathematics, to make proof more intuitive for non-specialists.

According to Hofstadter, two key concepts underlie Gödel's proof:

1) TNT contains strings that refer to other TNT strings, meaning TNT possesses self-reflection as a language.

2) Self-reflection can be concentrated into a single string, where the only focus of that string is itself. This idea stems from Cantor's diagonal argument (Hofstadter 1999).

[15] GEB also explores 'Strange Loops' through Escher and Bach, conceptualizing their structure as an 'Eternal Golden Braid.' However, the core mechanism of the 'Strange Loop' is fundamentally based on Gödel's incompleteness theorem.

This is followed by generating a new branch in number theory by incorporating either $G'$ or $\sim G''$ into TNT + $G$, leading to the axiomatic expansion of the formal system through an additional axiom schema:

$$(TNT+G_\omega) \qquad (12)$$

[Finally, Hofstadter suggests the following:] If there is a way of capturing the various strings $G$, $G'$, $G''$, $G'''$, ... in a single typographical mold, then there is a way of describing their Gödel numbers in a single arithmetical mold. And this arithmetical portrayal of an infinite class of numbers can then be represented inside TNT + $G_\omega$ by some formula OMEGA − AXIOM$\{a\}$ whose interpretation is: " $a$ is the Gödel number of one of the axioms coming from $G_\omega$". When $a$ is replaced by any specific numeral, the formula which results will be a theorem of TNT + $G_\omega$ if and only if the numeral stands for the Gödel number of an axiom coming from the schema. With the aid of this new formula, it becomes possible to represent even such a complicated notion as TNT + $G_\omega$-Proof-Pairs inside TNT + $G_\omega$ (Hofstadter 1999, p468):

$$(TNT+G_\omega)-PROOF-PAIR\{a, a'\} \qquad (13)$$

## B. Characteristics of "(TNT+$G_\omega$)−PROOF−PAIR{a, a'}" and Escher's "Dragon Biting Its Tail"

[Meanwhile, in Hofstadter's reformulated theorem, a question may arise:] "Why isn't $G_{\omega+1}$ among the axioms created by the axiom schema $G_\omega$?"

[In response to this, Hofstadter explains that such a paradox emerges in a way similar] to the Gödel's trick, in which the system's own properties are reflected inside the notion of proof pairs, and then used against the system. [In other words,] Any system, no matter how complex or tricky it is, can be Gödel-numbered, and then the notion of its proof-pairs can be defined and this is the petard by which it is hoist. Once a system is well-defined, or "boxed", it becomes vulnerable (Hofstadter 1999, p468-469).[16]

In summary, once this ability for self-reference is attained, the system has a hole which is tailor-made for itself; the hole takes the features of the system into account and uses them against the system

---

[16] The process by which every system is assigned a Gödel number and subsequently establishes the concept of proof pairs is only possible when a duplex relationship is formed between the 'content' that is to be proven and the 'formal structure' that seeks to prove it. However, as soon as this duplex relationship is established, as will be discussed later, the system inherently develops a paradoxical vulnerability - that is, however, this inevitably leads to a self-referential paradox akin to being trapped by one's own construction

(Hofstadter 1999, p471). That is, no matter how systematically all $G$ elements are embedded into TNT in a well-defined manner, an unforeseen and unaccounted for new $G$ will always emerge beyond the axiom schema.

If viewed solely from within the system, this paradoxical relationship can only be concluded as a self-destructive incompleteness. Even when examined from the perspective of completeness, Hofstadter's reformulation of TNT does not escape the self-destructive conclusion of incompleteness. This is because incompleteness is an intrinsic characteristic of TNT itself. As previously discussed, mathematics, as a tool constructed upon a closed-world assumption, pursues uniqueness and completeness while generating its own axioms. Given this nature, this paradox is unavoidable. Based on this reasoning, (Hofstadter 1999) attempts to elevate his reinterpretation of Gödel's TNT to a higher dimension.

[To illustrate this, Hofstadter introduces Escher's artwork:] his drawing Dragon. Its most salient feature is, of course, its subject matter-a dragon biting its tail, with all the Gödelian connotations which that carries. But there is a deeper theme to this picture, Escher himself wrote the following most interesting comments. The first comment is about a set of his drawings all of which are concerned with "the conflict between the flat and the spatial"; the second comment is about Dragon in particular (Hofstadter 1999, p473).

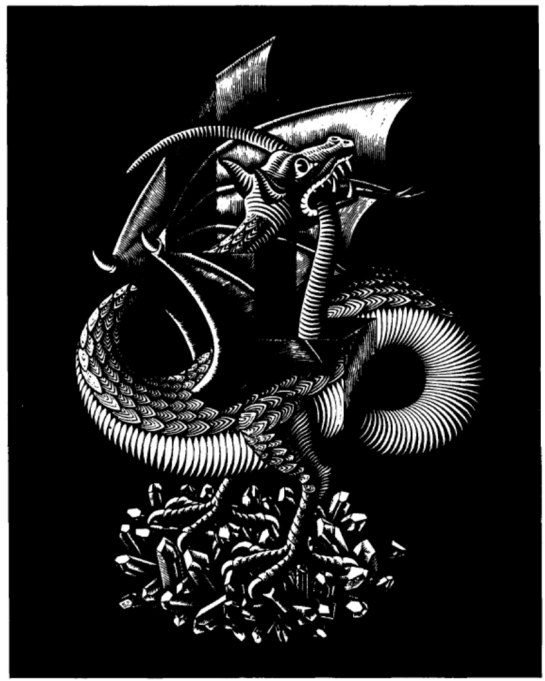

Figure 5. Dragon, by M. C. Escher (wood-engraving, 1952). The image is adapted from (Hofstadter 1999, p474, Figure 76).

I. Our three-dimensional space is the only true reality we know. The two dimensional is every

bit as fictitious as the four-dimensional, for nothing is flat, not even the most finely polished mirror. And yet we stick to the convention that a wall or a piece of paper is flat, and curiously enough, we still go on, as we have done since time immemorial, producing illusions of space on just such plane surfaces as these. Surely it is a bit absurd to draw a few lines and then claim: "This is a house". This odd situation is the theme of the next five pictures [including Dragon].

II. However much this dragon tries to be spatial, he remains completely flat. Two incisions are made in the paper on which he is printed. Then it is folded in such a way as to leave two square openings. But this dragon is an obstinate beast, and in spite of his two dimensions he persists in assuming that he has three; so he sticks his head through one of the holes and his tail through the other (M. C. Escher, The Graphic Work of M. C. Escher (New York: Meredith Press, 1967), p21-P22; Hofstadter 1999, p473).

[Hofstadter explains that Escher's intended] message is that no matter how cleverly you try to simulate three dimensions in two, you are always missing some "essence of three-dimensionality"(Hofstadter, 1999, p. 473). This insight underscores the idea that dimensional expansion cannot be fully captured by mere quantitative change but requires qualitative transformation. It also aligns with the previously stated conclusion that even the infinite expansion of TNT within the closed mathematical framework inevitably leads to incompleteness.

## C. "(TNT+$G_\omega$)–PROOF–PAIR{a, a'}" and DCPSs of TC-EO

Despite this, (Hofstadter 1999) reformulation of TNT presents its own intrinsic value. [According to (Hofstadter 1999), this new theorem[17]] brings up the fascinating concept of trying to create a computer program which can get outside of itself, see itself completely from the outside, and apply the Gödel zapping-trick to itself (Hofstadter 1999, p476). Based on this premise, this study posits that the core implication of Hofstadter's repeated application of Gödelian reasoning is that it leaves room for the idea of "Expansion of Dimensions as an Open World" to be incorporated into traditional Gödelian reasoning. By doing so, instead of merely assessing a closed world within the completeness based only on the lower dimension as a closed world, we can open the way to dynamically approach the world that expands itself based on the higher dimension as an open world, even if it cannot be completely defined.

### a. Dimensional Expansion as an Open-World System and DCPSs of TC-EO

In this respect, this study suggests that "dimensional expansion as an open-world system" corresponds to the "dimensional expansion accompanied by qualitative transformation" proposed in CFP theory. Furthermore, the conceptual hypothesis that enables such an expansion is DCPSs of TC-EO. Since the DCPSs of TC-EO focuses on the dynamic expansion of the world as an open world based on the higher dimension, regardless of whether the axiom system is complete or not, it allows the Paradoxical - Stratified relationship that goes beyond the Duplex - Contradictory relationship, which is found in Hofstadter's new formula. In other words, for this transformation from a Duplex-Contradictory relationship to a Paradoxical-Stratified relationship, a fundamental shift in worldview is required. Unlike the closed-world mathematical axiom systems, DNNs as data science operate under an open-world assumption, inherently making such transformations possible.

---

[17] More precisely, (Hofstadter 1999) discussed this within other refutations against Lucas that asserted the baffling repeatability of the Gödel argument.

This raises an essential question: What differentiates data science, which operates as an open-world system, from the closed mathematical axiom system?

Gödel's proof, grounded in a mathematical axiomatic world, relies on recursion as its fundamental mechanism to evaluate the system's completeness as the primary criterion. In contrast, data science, functioning as an open-world system, is understood as a continuous process of evolution and creation driven by the interrelations among data. Since it is fundamentally rooted in traces of life (data) that ceaselessly interact and transform, the essential criterion for judgment in data science is the generation of dynamism, rather than completeness. Consequently, the interrelations among data, features within data, and interactions between data and algorithms can all be examined through the common perspective of dynamism-creation.

**b. Recursion and Dynamism-Creation in Data Science**

Another critical point to consider is that even in data existence, the process of dynamism-creation requires a recursive mechanism.

This notion is strikingly similar to economic-social relationships. For instance, in economic interactions, a rational agent typically calculates gains and losses based on independent self-interest through arithmetic assessments of give-and-take transactions. However, an alternative approach exists: by prioritizing concessions and sacrifices to achieve collective benefit, one can generate synergistic effects that are otherwise unattainable.[18] Consider an economic agent who, in pursuit of common interests, engages in self-sacrificial behavior, treating the counterparty's benefit as their own. This approach, in effect, mirrors recursion in relational existence. Moreover, the unexpected benefits derived from this synergistic approach foster a sense of communal unity, which aligns with the conceptual models of crSTR and ABICND. While this may appear as an incomplete, disadvantageous approach in the short term, the long-term communal synergy effect surpasses such limitations.

From this perspective, an economic agent's actions transcend contradiction, yielding a paradoxical outcome, which in turn enables the formation of a new stratified entity—a community. Thus, in relational existence, a process that enables self-sacrifice and concession—instead of mere arithmetic evaluations of mutual interest—bears structural similarities to recursion in TNT. Given this, it becomes necessary to formally define this process as a "Duplex-Contradictory-Paradoxical-Stratified structure (DCPSs)." Since the transformation from duplex-contradictory relationships to paradoxical-stratified relationships inherently requires recursion as a developmental mechanism, such transformations cannot be fully realized within a conventional plane based on Euclidean space.

---

[18] For a more detailed discussion, refer to the author's social science paper, "The Convergent Paradigm – The SS Principle for a New Economic Philosophy as Public Philosophy(forthcoming)."

Moreover, for the duplex-contradictory relationship to evolve into a paradoxical-stratified relationship, the recursive process must involve an intermediate phase of "Thorough Closure (Solitude) – Eternal Opening (TC-EO)." This process does not manifest actively but instead unfolds passively over time, revealing its results only in retrospect. Thorough Closure (TC) [or Thorough Solitude (TS)][19] appears in Gödelian reasoning as the system's inherent incompleteness. Meanwhile, the tendency toward Eternal Openness (EO) appears, in the closed-world mathematical framework, as a contradiction. However, in an open-world perspective, this very tendency beyond incompleteness serves as the driving force enabling the transition to higher dimensions—though this transformation always remains hidden and only manifests through retrospective outcomes.

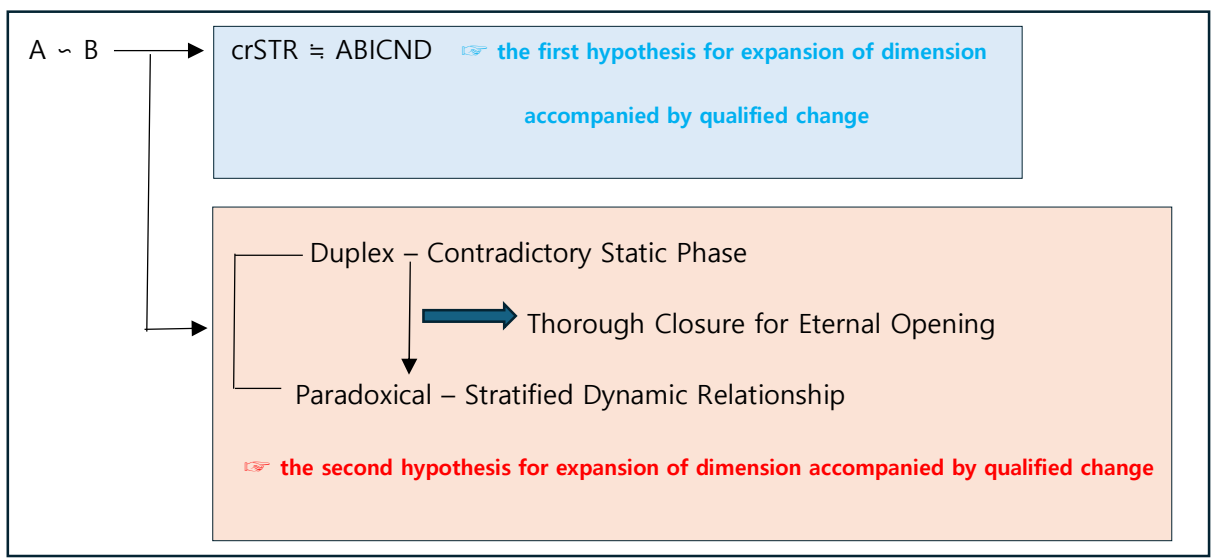

Table 2. New Conceptual hypothesis for dimensional expansion II

## 5) Summary

---

[19] In terms of adopting a proper term, in the field of data science involved in natural science, Thorough Closure (TC) may be more appropriate, while, in the field of humanities or social science such as economics, Thorough Solitude (TS) is more suitable.

This study has explored how the unique data-algorithm relationship modeling in DNNs creates a new developmental phase characterized by dimensional expansion accompanied by qualitative transformation. To articulate this idea, two conceptual hypotheses were proposed, drawing from existing achievements in physics and biology, as well as (Hofstadter 1999) reformulated theorem on Gödel's proof, namely (TNT+$G_\omega$)–PROOF–PAIR{a, a'}. Accordingly, this study defines this theoretical framework as the Convergent Fusion Paradigm (CFP theory), which conceptualizes a paradigm of data integration and convergence.

## IV. Application of CFP Theory

### 1. Mathematical and Technical Perspectives

### – Providing Philosophical and Theoretical Methodological Insights into the Geometric Structure of Data

The CFP theory, which integrates the two conceptual hypotheses discussed earlier, offers a new mathematical and technical perspectives for data geometry. By providing new insights into the relationships and structural combinations among data features, data points, and data-algorithm interactions, it establishes a structural framework that allows for the convergence and integration of high-dimensional and low-dimensional data structures, thereby more precisely capturing the data generation process.

Furthermore, by applying this structural framework to latent variable generative models such as VAEs and GANs, it enables the extraction of high-level information from high-dimensional data while preserving the structural integrity between low- and high-dimensional data representations.

In this context, this section examines the philosophical and theoretical methodological insights into data geometry, which can be obtained by applying CFP theory to the theoretical frameworks proposed by (Joshi et al. 2019) and (Arvanitidis et al. 2018, 2020).

### 1) (Joshi et al. 2019)

As previously mentioned, (Joshi et al. 2019) modifies the fundamental geometry to compute distances between observed feature values and decision boundaries. In other words, it treats the underlying quantitative vector spaces within Euclidean space as a virtual submanifold and calculates distances through an optimization process. However, this optimization process separates dimensional expansion from feature interrelations, contradicting the topological premise that features must necessarily exist within dimensional space.

By contrast, CFP theory does not separate dimensional expansion from feature interrelations. Instead, it interprets high-level features as emergent features, which are newly created through the relationships among low-level features. Furthermore, these emergent features themselves embody dimensional expansion accompanied by qualitative transformation. This structural development cannot be fully captured in Euclidean space, but it can be represented schematically as in Figure 5. Specifically, in Figure 5, as low-level features generate high-level features, the intermediate space (marked in green in Figure 5) represents a region that is difficult to capture under a Euclidean framework done by reductionist. This can be seen as another representation of the activation of hidden units in DNNs.

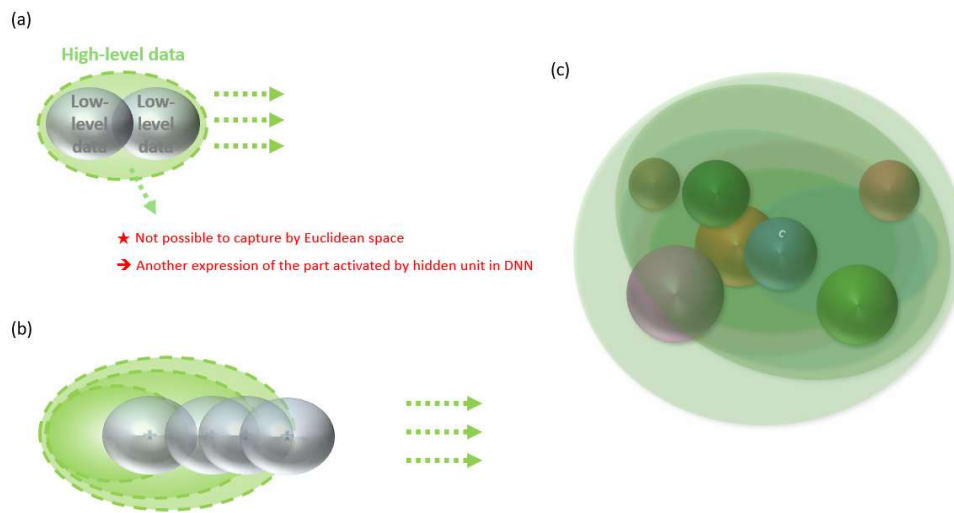

Figure 5. The Model of CFP Theory about Dimensional Expansion by the Relationships between Data Features. (a) The relationship between low- and high-level data features from the perspective of dimensional expansion accompanied by qualitative transformation (b) The relationship between low- and high-level data features from a fixed perspective (from one scale to another) (c) The relationship between low- and high-level data features from an overall perspective (from one scale to another).

**2) Data Manifold Hypothesis and Generative Model**

The CFP theory, which does not separate dimensional expansion from feature interrelations, becomes applicable when combined with the Riemannian data manifold hypothesis and generative models. Originally, a manifold is a mathematical concept referring to a topological space that locally resembles

Euclidean space but may exhibit complex global structures such as curvature (Kelley 2017).

The data manifold hypothesis extends this concept to data representation, suggesting that most naturally occurring high-dimensional datasets actually reside near a low-dimensional nonlinear manifold (Belkin and Niyogi 2003; Dominguez-Olmedo et al. 2023; Hastie and Stuetzle 1989; Smola et al. 2001). This hypothesis plays a crucial role in machine learning and data analysis, as it serves as a fundamental framework for understanding and utilizing the structural properties of data. Given that the data manifold hypothesis provides a geometric foundation to express the relationships between high- and low-dimensional data, it simultaneously offers a basis for applying CFP theory.

On the other hand, generative models offer an appealing framework to approximately learn the data manifold through differential geometric studies (Arvanitidis et al. 2018; Dominguez-Olmedo et al. 2023). In particular, the method of equipping a generative model with the Riemannian metric (Arvanitidis et al. 2018, 2019, 2020, 2021; Tosi et al. 2014) has addressed the inherent identifiability problem in conventional generative models, offering a more efficient method for learning data manifolds.[20]

When the two conceptual hypotheses of CFP theory are combined with the Riemannian manifold hypothesis under the conditions of generative models, it enables a more organic representation of the appearance with the embeddedly structured low-dimensional nonlinear manifold in the high-dimensional feature space. This enhanced structural expression allows for a more precise capture of the data generation process. Handling dimensional expansion between high- and low-dimensional data within Euclidean space by means of mathematical tools is inevitably limited to a quantitative approach. However, CFP theory facilitates dimensional expansion accompanied by qualitative transformation, based on its two conceptual hypotheses. Through this qualitative dimensional expansion process, low- and high-dimensional data structures can form a convergent and integrative structural relationship. In summary, CFP theory offers a geometric framework for expressing the convergent and integrative structural relationships between low- and high-dimensional data, which are difficult to capture within Euclidean space. This enables new insights into the data generation process.[21]

### 3) (Arvanitidis et al. 2018, 2020)[22]

Applying CFP theory to (Arvanitidis et al. 2018, 2020) provides philosophical and theoretical methodological insights into the geometric structure of data, particularly in relation to Riemannian

---

[20] Refer to II.1.3) b. for further discussion on this topic.

[21] For a more detailed explanation, refer to the author's paper, "A Novel Approach to Data Generation Processes within Generative Models" (Kim and Shim 2025).

[22] For a more detailed explanation, refer to the author's paper, "A Novel Approach to Data Generation Processes within Generative Models" (Kim and Shim 2025).

manifolds. These insights can be categorized into five key aspects:

### A. Redefining Riemannian Metrics in the Context of Dynamism

CFP theory reinterprets the Riemannian metric, traditionally regarded as a static and isolated mathematical unit, as a dynamic relational unit accompanied by dimensional transformation, viewed from a philosophical and ontological perspective.

### B. Philosophical and Logical Analysis of the Inverse Function of the Jacobian

From the perspective of CFP theory, using the inverse function of the Jacobian implies inserting reversed time into the Riemannian metric. This enables the Riemannian manifold hypothesis to express not merely quantitative changes in dimensions but qualitative transformations in relative space-time. It extends beyond a compression-restoration viewpoint toward a generative perspective in data structure formation in VAEs or GANs. Through this philosophical and theoretical understanding of the inverse function of the Jacobian, CFP theory explains how pull-back metrics facilitate the generation of structured relationships between high- and low-dimensional data representations.

### C. Understanding the Indirect Effect of Posterior Distributions in Generative Models

CFP theory provides insight into why the process of bringing a Riemannian metric from the ambient (input) space to the latent space *via* a pull-back metric produces an effect similar to utilizing posterior distributions $p(z|x)$. This phenomenon, referred to as the indirect effect of posterior distributions, offers a deeper understanding of latent space structure in generative models.

### D. Understanding Curvature, Data Density, and Learning in the Riemannian Ambient Metric

Understanding how the Riemannian ambient metric measures data density and curvature, in terms of the learning process, provides key insights into the geometric structure of data. From the perspective of CFP theory, the function of RBF networks in the ambient (input) space can be interpreted as a microscopic continuous developmental process featuring as a chain of crSTR.

### E. Interpreting Generators in Generative Models through CFP Theory

From the viewpoint of CFP theory, a generator can be understood as a mechanism that integrates the ambient (input) space and the latent space through a pull-back metric, ultimately incorporating them into the ambient (output) space.

## 2. Causal Perspective – Exploring New Possibilities of Abduction

If CFP theory can serve as a theoretical foundation for a new causal model that overcomes the challenge of the common principle problem inherent in causal approach, beyond just mathematical and technological perspectives mentioned above, then it is necessary to consider the potential of abduction, which differs from traditional deduction and induction,[23] as a methodological means for developing the specific causal model, aside from accepting backward causation within Hilbert space. This study argues that abduction, when utilizing Hilbert space, can serve as a fundamental methodological approach for constructing a new specific causal model.

## 1) Abduction as a Method of Investigating Causality

### A. Concept of Abduction and Peirce's Early Abduction as Syllogistic Form

To understand abduction as a logical methodology for investigating causality, it is first necessary to clarify its conceptual foundation. This requires a brief discussion of deduction and induction, which are the two primary modes of reasoning:

○ Deduction derives logically necessary specific propositions from universally validated general premises (Kaufmann 1999).

○ Induction derives general propositions based on commonalities observed in specific instances (Kaufmann 1999).

Thus, deduction follows a "general-to-specific" reasoning direction, while induction follows a "specific-to-general" approach, making them contrasting logical methods. Despite this difference, both deduction and induction share a common characteristic: they employ a vertical (hierarchical) logical structure.

However, abduction takes a different approach—instead of moving from general to specific (deduction) or from specific to general (induction), it traces backward from a given event to explore the premises upon which that event is based (Walton 2005). In other words, abduction is a form of hypothetical reasoning that seeks plausible premises to provide a reasonable explanation for an atypical case (Thagard 1988). Due to this characteristic, abduction is sometimes referred to as "inference to the best explanation" (Walton 2005).

---

[23] Deduction, induction, and abduction do not constitute causality itself. However, they serve as specific logical methodologies used to explore and understand causality, making them essential in causal investigations (Larson 2021).

While the notion of "premises upon which an event is based" and "the best possible explanation" might suggest a reasoning process similar to induction, the "general" inferred through abduction is not necessarily constrained to a vertical logical structure but is instead open to a horizontal logical framework. Thus, unlike deduction and induction, which predominantly rely on vertical logical structures, abduction does not completely reject vertical logic but is fundamentally characterized by a horizontal logical structure.

**B. Abduction as Logical Methodology in Peirce and Walton**

When examining abduction as a logical methodology, it is essential to recognize that it was formally developed by Charles Sanders Peirce (1839–1914), a 19th-century American philosopher and pragmatist. Peirce introduced abduction as a third mode of reasoning, distinct from deduction and induction, and argued that abduction is the only form of inference that can genuinely expand the scope of knowledge. Initially, Peirce conceptualized abduction as a process of forming hypotheses, describing it as an inverted version of the syllogistic form of deduction, which follows the pattern:

○ Deductive Syllogism: Major premise → Minor premise → Conclusion
○ Abductive Inference: Conclusion → Major premise → Minor premise

Peirce viewed abduction as a method for discovering hypotheses, rather than simply deducing conclusions from premises (Hartshorne et al. 1931).

However, reducing abduction to a mere inversion of the deductive syllogism oversimplifies its original intent, which is to serve as "inference to the best explanation" (Walton 2005). This approach confines the rich reasoning process of abduction within the rigid vertical structure of syllogistic logic, which fundamentally contradicts the nature of abductive reasoning. Most importantly, the process of formulating a hypothesis as the "best explanation" inherently involves epistemic shifts and expansions, which necessitate a departure from vertical structures toward a horizontal logical framework. Peirce himself, in his later works, acknowledged that abduction could not be fully captured by a syllogistic structure[24] and presented a non-syllogistic logical process as follows (Hartshorne et al. 1931):

---

[24] After 1898, Peirce explicitly distinguished his earlier theory as "hypothetic inference" or "qualitative induction", while referring to his later theory as "abductive inference" or simply "abduction" (Reichertz 2013).

① A surprising phenomenon *C* is observed.

② However, if hypothesis *A* were true, phenomenon *C* would be explained.

③ Therefore, there is reason to tentatively consider hypothesis *A* as true.

The most distinctive feature of this horizontal reasoning process is that, even if phenomenon *C* is observed, there is no logical necessity ensuring the truth of hypothesis *A*. Due to this characteristic, deductive reasoning inherently possesses logical necessity, while inductive reasoning approximates logical necessity by generalizing from observed cases. In contrast, abduction lacks logical necessity, leading many to question its validity as a rigorous methodological approach.

However, when abduction is placed within a dynamic, continuous reasoning process,[25] it can function as a progressive inferential method. Through the iterative elimination of irrational or irrelevant hypotheses, abduction moves toward a more plausible and acceptable conclusion. In other words, if hypothesis *A* best explains phenomenon *C*, and no alternative hypothesis provides a better explanation, then hypothesis *A* can be tentatively accepted as the most plausible explanation (Bongiovanni et al. 2018). Since hypothetical reasoning within a horizontal logical structure operates dynamically, abduction gradually converges on the most plausible explanation, much like inductive reasoning approximates logical necessity. Recognizing this, (Walton 2002) reformulated abduction as follows:

① *F* represents the discovered or given facts.

② *E* is a satisfactory explanation of *F*.

③ No alternative explanation *E′* is as satisfactory as *E*.

④ Therefore, *E* is a plausible hypothesis.

**2) Application of CFP Theory**

**A. Understanding Abduction Based on Critical Realist Ontology**

---

[25] Due to this progressive process, abduction is conceptually closely related to retroduction. However, the relationship between abduction and retroduction has been extensively debated, leading to various interpretations [(Chiasson 2005; Downward and Mearman 2007; Hanson 1979; Magnani 1999; Rescher 1978; Skagestad 1981)]. Since this study focuses on abduction, a detailed discussion of retroduction falls outside the scope of this paper.

(Wuisman 2005) attempts to understand abduction from the perspective of the critical realist ontology. According to (Bhaskar 1998; Wuisman 2005, p51), the critical realist approach to social scientific research starts from the ontological notion that social reality is stratified. In critical realist literature three hierarchically arranged layers are distinguished: the empirical, the actual and the real, at which experiences, events and mechanisms are, respectively, situated. When analyzing these three layers in light of Walton's reformulation of abduction (Walton 2002), experience corresponds to $F$ (discovered or given facts), event corresponds to $E$ (a satisfactory explanation of $F$), and mechanism corresponds to the specifics of step ③ (no alternative explanation $E'$ is as satisfactory as $E$). [Based on this framework, critical realism regards abduction] as the modes of inference specifically required to explore underlying levels of reality and uncover their mechanisms and events (Wuisman 2005, p369).

Building upon this understanding, (Wuisman 2005) presents the logic of inquiry for abduction as follows:

[If analyzing the classical Greek statue under a critical realist approach, different from some cases such as] the Rejang example [in which] the logic of inquiry consisted of a combination of induction and deduction (Groot and A 2020), the gap between the appearances of the statue observable through the senses and the underlying mechanism, from which it is thought to have emerged, seems unbridgeable. What mode of inference or logic, if any, is capable of linking the sensory perceptions of the statue to this underlying mechanism? The mental acrobatics required for this seem to imply a creative leap, to say the least. That means quite clearly that induction and deduction are of no use. In the literature on the various modes of inference, the only kind of reasoning that is considered to involve a creative leap is called 'abduction' (Fischer 2001, p380-p381; Wuisman 2005, p361-p383).

Thus, from the perspective of Walton's reformulation of abduction (Walton 2002), this creative leap corresponds to step ④ (plausibility of hypothesis $E$ as an explanation for $F$). In summary, (Wuisman 2005) understands abduction as a reasoning process that moves from types of events captured at the empirical level through a creative leap toward the mechanisms at the level of reality, which are presumed to have triggered those experiences.

**B. crSTR as Mechanism and DCPSs of TC-EO as Creative Leap**

When this understanding of abduction, grounded in the critical realist ontology, is reinterpreted through the premises of Hilbert space, which incorporates backward causation and CFP theory, it provides a potential foundation for a new causal model.

First, when abduction is understood within the spatial framework of Hilbert space, its horizontal logical structure, simultaneously - to a degree - implicating vertical structures, can be spatially expanded and deepened through the development of dimensionality, as expressed through a manifold. In this context, Hilbert space is regarded as a background space to make it possible to accept abduction based on the critical realist approach as the creative leap. Next, based on this premise, when abduction, understood through the ontology of critical realism, is examined through CFP theory, the three hierarchical layers—experience, event, and mechanism—can be mapped as follows:

1. Experience ($F$) and Event ($E$) correspond to the two interacting entities in CFP theory.

2. Mechanism corresponds to crSTR, created through the convergence and fusion of these entities, as well as ABICND, a newly generated entity.

3. The creative leap required in abduction corresponds to DCPSs of TC-EO.

This interpretation enriches the causal understanding of emergent features expressed through backward causation in Hilbert space, providing a foundation for developing a concrete causal model.

## V. Conclusion

In summary, CFP theory has the potential to serve as the foundational framework for a new causal model capable of overcoming the fundamental challenge posed by common cause. It not only provides insights into the geometric structure of data based on Riemannian manifolds but also establishes the mathematical, technological, and causal-philosophical foundation necessary for elucidating the relationships between data features. This, in particular, contributes to a more precise positioning of causal inference within machine learning, especially in DNNs.

Meanwhile, it is crucial to examine the potential of Hilbert space concerning data generation. Unlike Euclidean space, Hilbert space is not merely a medium for representing data but is more suited for analyzing complex interactions and nonlinear relationships. However, it cannot be arbitrarily configured as an infinite, unstructured space, as it is constrained by the convergence conditions of Cauchy sequences and the limitations of linear matrix operations (Kim et al. 2025). Therefore, the data generation process within Hilbert space should not be regarded as an entirely novel nonlinear generation but rather as an approximation achieved through increasingly refined and expanded quantitative measurement scales. For instance, this can be likened to creating an animation by linking sequential images—where adding more frames allows for a smoother and more organic motion, but each individual frame itself does not become the animation. In other words, while Hilbert space can go beyond the 'Duplexity-Contradiction' structure of Euclidean space and accommodate a 'Paradoxical-Stratified' structure, it still faces limitations in fully realizing nonlinear interactions and qualitative generative processes.

Then, is it possible to develop a mathematical and technological methodology that directly represents qualitative data generation as an inherently nonlinear process, rather than relying on linear approximation techniques? Nonlinear interactions are closely related to the principles of superposition and entanglement observed in quantum mechanics. To realize such an approach, one might consider utilizing the Schrödinger equation, integrating general relativity with quantum mechanics, or leveraging quantum computing technologies. However, it remains difficult to assert that such methodologies would entirely depart from conventional linear approaches. Until now, qualitative generative processes that should inherently be nonlinear have been predominantly expressed through probabilistic approximations (e.g., Bayesian methods) and continuous, linear mathematical models (e.g., differential equations). Nevertheless, these remain fundamentally post hoc approximations, and no appropriate mathematical tool capable of directly capturing nonlinearity in its entirety has yet been proposed.

Given these practical limitations, this study emphasizes the necessity of supplementing existing mathematical and technological methodologies with an enhanced philosophical perspective. Specifically, there is a need for an in-depth examination of passivity, or the concept of "Leave As it Is (LAI)," as it exists within the DCPSs of TC-EO in CFP theory. This suggests that the position of a fully functional nonlinear qualitative generative process may be understood through the domain of passivity inherent in CFP theory's theoretical development.[26] In other words, this study posits that nonlinear generativity, which cannot be adequately captured by existing mathematical and technological methodologies alone, may be accessed through an understanding of passivity embedded within CFP theory's framework. This new structural approach, integrating philosophical insight, fundamentally differs from conventional probabilistic approximation techniques and has the potential to explore the generative mechanisms underlying quantum superposition and entanglement. Consequently, such an approach may offer novel insights into addressing fundamental and critical issues in LLMs, such as hallucination, from a completely different perspective.

---

[26] The area of passivity assumes that when thorough convergence surpasses a certain critical point, it can be expressed as a fusion that opens up a new dimension. In this context, assuming the fusion implies the operation of passivity. For a more detailed discussion, refer to (Kim and Shim 2025), *"A Novel Approach to Data Generation in Generative Models."*